\documentclass[sigconf]{acmart}
\AtBeginDocument{%
  }

\copyrightyear{2025}
\acmYear{2025}
\setcopyright{acmlicensed}\acmConference[MM '25]{Proceedings of the 33rd ACM International Conference on Multimedia}{October 27--31, 2025}{Dublin, Ireland}
\acmBooktitle{Proceedings of the 33rd ACM International Conference on Multimedia (MM '25), October 27--31, 2025, Dublin, Ireland}
\acmDOI{10.1145/3746027.3755168}
\acmISBN{979-8-4007-2035-2/2025/10}

\usepackage{graphicx}               
\usepackage{colortbl}   

\usepackage{subcaption}             
\usepackage{wrapfig}                
\usepackage{float}                  

\usepackage{amsmath}                 
\usepackage{amssymb}                 

\usepackage{bm}                       

\usepackage{booktabs}                
\usepackage{multirow}                
\usepackage{enumitem}                

\usepackage{algorithm}
\usepackage{algorithmic}

\usepackage{tcolorbox}               
\usepackage{ulem}                     
\usepackage{geometry}                 
\usepackage{pifont}                   
\usepackage{balance}                  

\usepackage{caption}                  
\usepackage{hyperref}                 

\begin{document}

\title{%
  ANT: Adaptive Neural Temporal-Aware Text-to-Motion Model%
}





\author{Wenshuo Chen}
\authornote{Represents equal contribution.}
\email{wenshuochen@hkust-gz.edu.cn}
\affiliation{%
  \institution{HKUST (GZ)}
  \city{Guangzhou}
  \country{China}
}

\author{Kuimou Yu}
\authornotemark[1]
\email{kyu745@connect.hkust-gz.edu.cn}
\affiliation{%
  \institution{HKUST (GZ)}
  \city{Guangzhou}
  \country{China}
}

\author{Jia Haozhe}
\authornotemark[1]
\email{haozhejia@hkust-gz.edu.cn}
\affiliation{%
  \institution{HKUST (GZ)}
  \city{Guangzhou}
  \country{China}
}

\author{Kaishen Yuan}
\email{kaishenyuan@hkust-gz.edu.cn}
\affiliation{%
  \institution{HKUST (GZ)}
  \city{Guangzhou}
  \country{China}
}

\author{Zexu Huang}
\email{Zexu.Huang@student.uts.edu.au}
\affiliation{%
  \institution{University of Technology Sydney}
  \department{School of Electrical and Data Engineering (SEDE)}
  \city{Sydney}
  \state{NSW}
  \country{Australia}
}

\author{Bowen Tian}
\email{bowentian@hkust-gz.edu.cn}
\affiliation{%
  \institution{HKUST (GZ)}
  \city{Guangzhou}
  \country{China}
}

\author{Songning Lai}
\email{songninglai@hkust-gz.edu.cn}
\affiliation{%
  \institution{HKUST (GZ)}
  \city{Guangzhou}
  \country{China}
}

\author{Hongru Xiao}
\email{hongru_xiao@tongji.edu.cn}
\affiliation{%
  \institution{Tongji University}
  \department{College of Civil Engineering}
  \city{Shanghai}
  \country{China}
}

\author{Erhang Zhang}
\email{seve19861@gmail.com}
\affiliation{%
  \institution{Shandong University}
  \department{Chongxin}
  \city{Qingdao}
  \country{China}
}

\author{Lei Wang}
\email{l.wang4@griffith.edu.au}
\affiliation{%
  \institution{Griffith University}
  \city{Brisbane}
  \state{Queensland}
  \country{Australia}
}
\affiliation{%
  \institution{Data61/CSIRO}
  \city{Canberra}
  \state{ACT}
  \country{Australia}
}

\author{Yutao Yue}
\authornote{Correspondence to Yutao Yue <yutaoyue@hkust-gz.edu.cn>}
\email{yutaoyue@hkust-gz.edu.cn}
\affiliation{%
  \institution{HKUST (GZ)}
  \department{Thrust of Artificial Intelligence and Thrust of Intelligent Transportation}
  \city{Guangzhou}
  \country{China}
}

\settopmatter{authorsperrow=4}





\renewcommand{\shortauthors}{Wenshuo Chen et al.}

\begin{abstract}

While diffusion models advance text-to-motion generation, their static semantic conditioning ignores temporal-frequency demands: early denoising requires structural semantics for motion foundations while later stages need localized details for text alignment. This mismatch mirrors biological morphogenesis where developmental phases demand distinct genetic programs. Inspired by epigenetic regulation governing morphological specialization, we propose \textbf{\underline{(ANT)}}, an \textbf{\underline{A}}daptive \textbf{\underline{N}}eural \textbf{\underline{T}}emporal-Aware architecture. ANT orchestrates semantic granularity through: \textbf{\underline{(i)} Semantic Temporally Adaptive (STA) Module:} Automatically partitions denoising into low-frequency structural planning and high-frequency refinement via spectral analysis. \textbf{\underline{(ii)} Dynamic Classifier-Free Guidance scheduling (DCFG):} Adaptively adjusts conditional to unconditional ratio enhancing efficiency while maintaining fidelity. Extensive experiments show that ANT can be applied to various baselines, significantly improving model performance, and achieving state-of-the-art semantic alignment on StableMoFusion. Code can be found on \href{https://github.com/CCSCovenant/ANT}{https://github.com/CCSCovenant/ANT}. 
\begin{flushright}
\textit{"Details make perfection, and perfection is not a detail."} \\  
— Leonardo da Vinci  
\end{flushright}
\end{abstract}



\begin{teaserfigure}
    \vspace{-15pt}
    \centering
  \includegraphics[width=\textwidth]{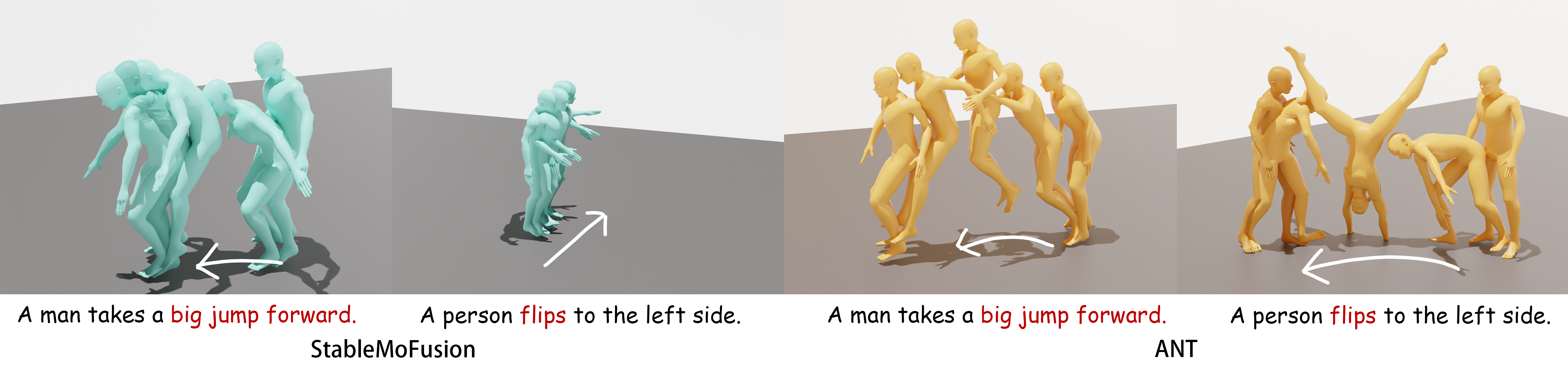}
  \captionsetup{skip=2pt}
  \caption{Our ANT can be seamlessly plugged into diffusion-based text-to-motion models to generate semantically rich, fine-grained, and naturally smooth motions with high precision.}
  \Description{Enjoying the baseball game from the third-base
  seats. Ichiro Suzuki preparing to bat.}
  \label{fig:teaser}
\end{teaserfigure}

\maketitle

\section{Introduction}

Text-driven human motion generation has recently attracted significant attention due to the semantic richness and intuitive nature of natural language descriptions, with broad applications in animation, film production, virtual/augmented reality (VR/AR), and robotics \cite{10.1145/3664647.3681034,chen2025freet2mfrequencyenhancedtexttomotion, pinyoanuntapong2024mmmgenerativemaskedmotion, controlmm,bamm}. While textual prompts provide valuable semantic guidance for motion synthesis, they often suffer from incomplete or inaccurate semantic representations, leading to suboptimal generation quality \cite{10.1145/3664647.3681034,tevet2022humanmotiondiffusionmodel}. Ensuring faithful alignment between generated motions and textual descriptions remains a critical challenge.

Current research in text-to-motion generation primarily focuses on two paradigms: VAE-based models that encode motions into discrete tokens for prediction using autoregressive (AR) \cite{t2mgpt,jiang2023motiongpthumanmotionforeign} or non-autoregressive (NAR) \cite{guo2023momask,pinyoanuntapong2024mmmgenerativemaskedmotion,bamm} frameworks, and diffusion-based models that gradually transform Gaussian noise into motion sequences through iterative denoising under text conditioning \cite{mdm, chen2023executingcommandsmotiondiffusion, huang2024stablemofusionrobustefficientdiffusionbased,zhang2023remodiffuse,dai2024motionlcmrealtimecontrollablemotion}. Among these approaches, vector quantization (VQ)-based discrete generation methods have become the dominant paradigm in human motion synthesis \cite{guo2023momask,bamm}. However, despite their effectiveness, these methods suffer from inherent drawbacks, including information loss and reduced motion diversity \cite{t2mgpt, guo2022tm2tstochastictokenizedmodeling}. Conversely, diffusion-based approaches offer unique advantages, such as fine-grained detail generation, diverse motion outputs, and physically plausible movement synthesis, making them a promising alternative \cite{yuan2023physdiffphysicsguidedhumanmotion, zhang2022motiondiffusetextdrivenhumanmotion,mdm,chen2025freet2mfrequencyenhancedtexttomotion}. Nevertheless, diffusion-based methods still lag behind VQ-based models in terms of overall performance \cite{huang2024stablemofusionrobustefficientdiffusionbased, mdm}.

Recent efforts have sought to bridge this performance gap by enhancing motion representations within diffusion models \cite{chen2025freet2mfrequencyenhancedtexttomotion, hong2025saladskeletonawarelatentdiffusion}. One notable approach involves projecting motion data into a compact and fine-grained latent space using a 1D ResNet-based autoencoder, thereby improving motion structure modeling and prediction accuracy \cite{hong2025saladskeletonawarelatentdiffusion}. While such methods mitigate some of the limitations of VAE-based approaches, they remain significantly inferior to state-of-the-art techniques. Moreover, recent studies \cite{10.1145/3664647.3681034} reveal that diffusion-based text-to-motion models exhibit limitations in aligning generated motions faithfully with input text descriptions. To better understand the generative process of diffusion-based models, prior work \cite{chen2025freet2mfrequencyenhancedtexttomotion} has analyzed the denoising mechanism and proposed a two-phase generation framework: a semantic planning stage for low-frequency feature modeling and a fine-grained refinement stage for high-frequency generating. This decomposition raises a crucial research question: \textbf{What distinct roles does textual information play in these two phases? How can we enhance the semantic alignment of diffusion-based approaches to achieve more accurate and expressive motion synthesis?}

Drawing inspiration from biological processes, Hinton likened the metamorphosis of insects to different stages of computational learning: larval stages prioritize energy absorption, while adult stages focus on locomotion and reproduction \cite{hinton2015distillingknowledgeneuralnetwork}.
Analogously, the diffusion process in motion generation should follow a phase-wise prioritization, capturing low-frequency motion structures in early denoising steps and refining high-frequency motion details in later stages. However, existing methods \cite{huang2024stablemofusionrobustefficientdiffusionbased, mdm, chen2025freet2mfrequencyenhancedtexttomotion} apply CLIP-encoded \cite{clip} text embeddings uniformly across all denoising steps, failing to distinguish between these two phases. This uniform application can lead to incomplete motion structures in early stages and insufficient detail refinement in later stages. However, explicitly decomposing semantic information in the frequency domain remains a significant challenge.

To address these issues, we propose ANT (Adaptive Neural Temporal-Aware Text-to-Motion Model). Unlike conventional diffusion based approaches, ANT incorporates a plug-and-play adaptive Semantic Temporal-Aware Module (STA) and a Dynamic schedule method (DCFG) based CFG \cite{ho2022classifierfreediffusionguidance}. STA model dynamically adjusts its response at different timesteps in the denoising process. This module adapts autonomously without requiring explicit supervision, enabling progressive semantic modulation. Specifically, STA prioritizes global motion structures (low-frequency) during the early denoising steps and refines detailed motion variations (high-frequency) in the later stages. For the CFG Schedule, based on STA's ability to distinguish between early and late stages in text attention, we dynamically adjust the ratio between conditional and unconditional results during sampling \cite{lu2022dpmsolverfastodesolver,song2022denoisingdiffusionimplicitmodels}. In the later stages, we switch to the more efficient unconditional generation. Through these temporally-aware method, our method improves semantic alignment, resulting in more accurate, diverse, and physically plausible motion synthesis.
\begin{figure}[t]
    \centering
    \includegraphics[width=\columnwidth]{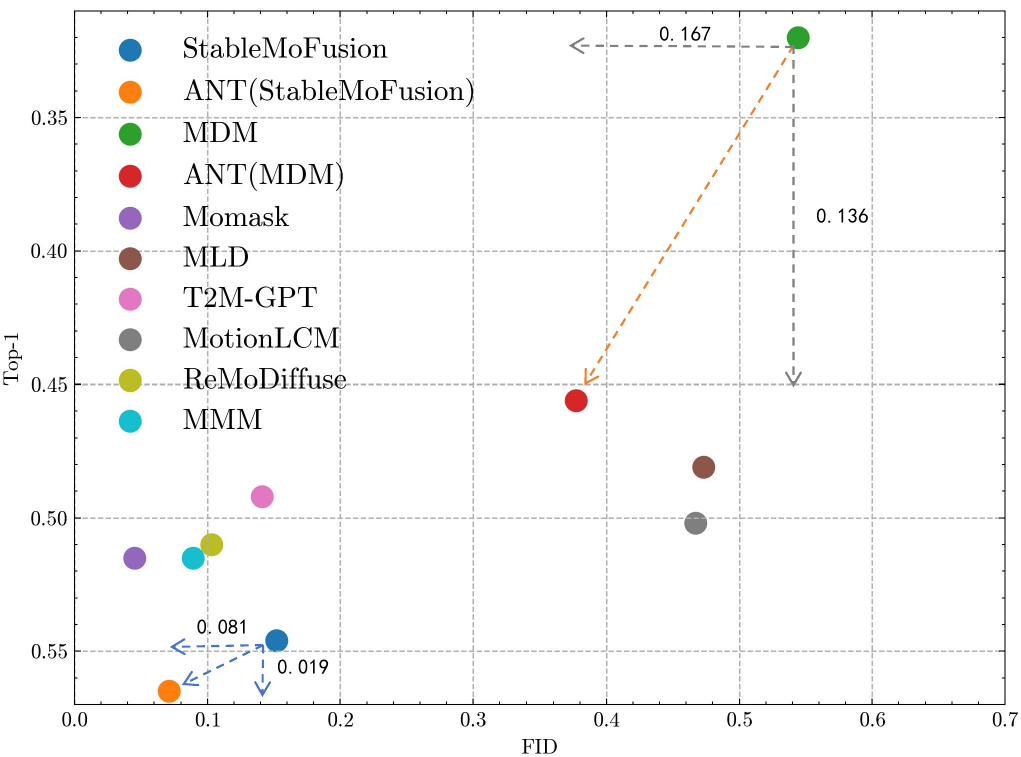}
    \vspace{-15pt}
    \caption{Comparison with SOTA models. The figure presents a comparison with the best-performing text-to-motion models to date, where a closer distance to the origin indicates better overall performance. ANT achieves significant improvements when applied to both MDM and StableMoFusion. Notably, ANT on StableMoFusion outperforms all other models in terms of R-Precision, highlighting its effectiveness and superiority among state-of-the-art methods.}
    \vspace{-0.6cm}
    \label{fig:plot}
\end{figure}
We conducted experiments on MDM \cite{mdm} and StableMoFusion \cite{huang2024stablemofusionrobustefficientdiffusionbased}. The results demonstrate significant improvements in FID and R-Precision. Additionally, by taking StableMoFusion as its baseline, ANT surpasses representative VAE-based models (Figure \ref{fig:plot}), such as MMM \cite{pinyoanuntapong2024mmmgenerativemaskedmotion} and T2M-GPT \cite{t2mgpt}, across all metrics, further demonstrating the potential of diffusion-based methods. Our method is simple and efficient, introducing minimal additional computational overhead while being effective across different architectures, including DDIM \cite{huang2024denoisingdiffusionprobabilisticmodels} and DPM-Solver \cite{lu2023dpmsolverfastsolverguided}. This provides a novel and interpretable research direction for diffusion-based text-to-motion methods.

We summarize our contributions as follows:
\begin{enumerate}
    \item We propose the first dynamic text encoding modulation framework for text-to-motion generation by analyzing the denoising mechanism. The integration of our Semantic Temporal Awareness Module significantly improves alignment between generated motions and textual descriptions.
    \item We conduct an in-depth analysis of the denoising mechanism in text-to-motion generation and provide a new DCFG schedule Method to improve sampling efficiency.
    \item Extensive experiments demonstrate that our method achieves state-of-the-art performance in text-motion alignment, validated by quantitative metrics and user studies.
\end{enumerate}

\section{Related Work}
\textbf{Text-to-Motion Generation.} Recent advancements in text conditioned human motion generation have been driven by two primary methodologies: diffusion-based models and VAE-based models. Diffusion models \cite{tevet2022humanmotiondiffusionmodel,tevet2022motionclipexposinghumanmotion,zhang2024motiondiffuse,chen2023executingcommandsmotiondiffusion,dai2024motionlcmrealtimecontrollablemotion,zhang2023remodiffuse}  have shown remarkable potential in modeling the complex relationships between textual inputs and motion sequences. Prominent works include MotionDiffuse \cite{zhang2024motiondiffuse}, which leverages cross-attention for text integration; MDM \cite{mdm}, which explores diverse denoising networks such as Transformer and GRU; PhysDiff \cite{yuan2023physdiff}, which incorporates physical constraints to enhance realism. While ReMoDiffuse \cite{zhang2023remodiffuse} improves performance through retrieval mechanisms, MotionLCM \cite{dai2024motionlcmrealtimecontrollablemotion} achieves real-time, controllable generation via a latent consistency model.

On the other hand, VAE-based models have also demonstrated strong performance in multi-modal motion generation. ACTOR \cite{petrovich2021actionconditioned3dhumanmotion} proposes a Transformer-based VAE for generating motion from predefined action categories, while TEMOS \cite{petrovich2022temosgeneratingdiversehuman} extends ACTOR with an additional text encoder to support diverse motion sequences, primarily focusing on short sentences. Guo et al. \cite{guo2022generating} introduce an autoregressive conditional VAE that conditions on generated frames and text features, predicting motions based on text length. TEACH \cite{athanasiou2022teachtemporalactioncomposition} builds upon TEMOS to enable the generation of longer, temporally coherent motion compositions from sequential natural language descriptions. TM2T \cite{guo2022tm2t} jointly trains both text-to-motion and motion-to-text tasks, improving bidirectional generation quality. T2M-GPT \cite{t2mgpt} quantizes motion clips into discrete markers and utilizes a transformer-based model to generate subsequent markers.

Despite the advancements in both diffusion-based and VAE-based approaches, a common limitation persists: reliance on the CLIP encoder \cite{clip}. Many existing methods, including diffusion-based models \cite{mdm,zhang2024motiondiffuse,huang2024stablemofusionrobustefficientdiffusionbased} and VAE-based models like MotionCLIP \cite{tevet2022motionclipexposinghumanmotion}, process textual descriptions through CLIP to obtain fixed text feature representations. However, this static encoding fails to provide rich, dynamic semantic information throughout the motion generation process. As a result, models struggle to adaptively interpret textual nuances over time, leading to inconsistencies in generated motions and limiting expressiveness.

\textbf{Senabtic embedding of Text-to-Motion}In the field of text-to-motion synthesis, a prevalent strategy involves leveraging the CLIP text encoder \cite{clip} to derive semantic embeddings. This approach is adopted by models such as MoMask \cite{guo2023momask} and StableMoFusion \cite{huang2024stablemofusionrobustefficientdiffusionbased}. Alternative methodologies incorporate pretrained word embeddings in conjunction with sequence processing layers, typically Transformers or Gated Recurrent Units (GRUs) \cite{guo2022tm2t}, as demonstrated in the TM2T framework. More recent developments have seen the integration of novel encoders: MotionGPT \cite{jiang2023motiongpthumanmotionforeign} employs the T5 model, while MDM \cite{mdm}, initially utilizing CLIP, has subsequently experimented with BERT as an alternative text representation module.

However, each of these encoding strategies exhibits distinct limitations. Specifically, the GRU-based encoder architecture, as implemented in TM2T \cite{guo2022tm2t}, encounters difficulties with processing extended sequences, capturing long-range dependencies effectively, and maintaining computational efficiency. These shortcomings can potentially impair the nuanced understanding of complex textual prompts and consequently reduce the fidelity of the generated motion sequences. Although encoders predicated on CLIP, T5 \cite{raffel2023exploringlimitstransferlearning} and BERT \cite{devlin2019bertpretrainingdeepbidirectional} generally yield more robust textual representations, a significant constraint stems from the static nature of their output embeddings throughout the iterative diffusion denoising process. As these embeddings do not dynamically evolve in relation to the diffusion time step, they may not sufficiently address the global conditioning requirements intrinsic to sophisticated generative modeling paradigms. Contrasting existing techniques our method dynamically modulates text embeddings per timestep via a Semantic Temporal-Aware Module (STA). This module aligns adaptive semantic evolution with the denoising process boosting motion-text alignment and generation fidelity.

\begin{figure*}
    \centering
    \includegraphics[width=0.85\linewidth]{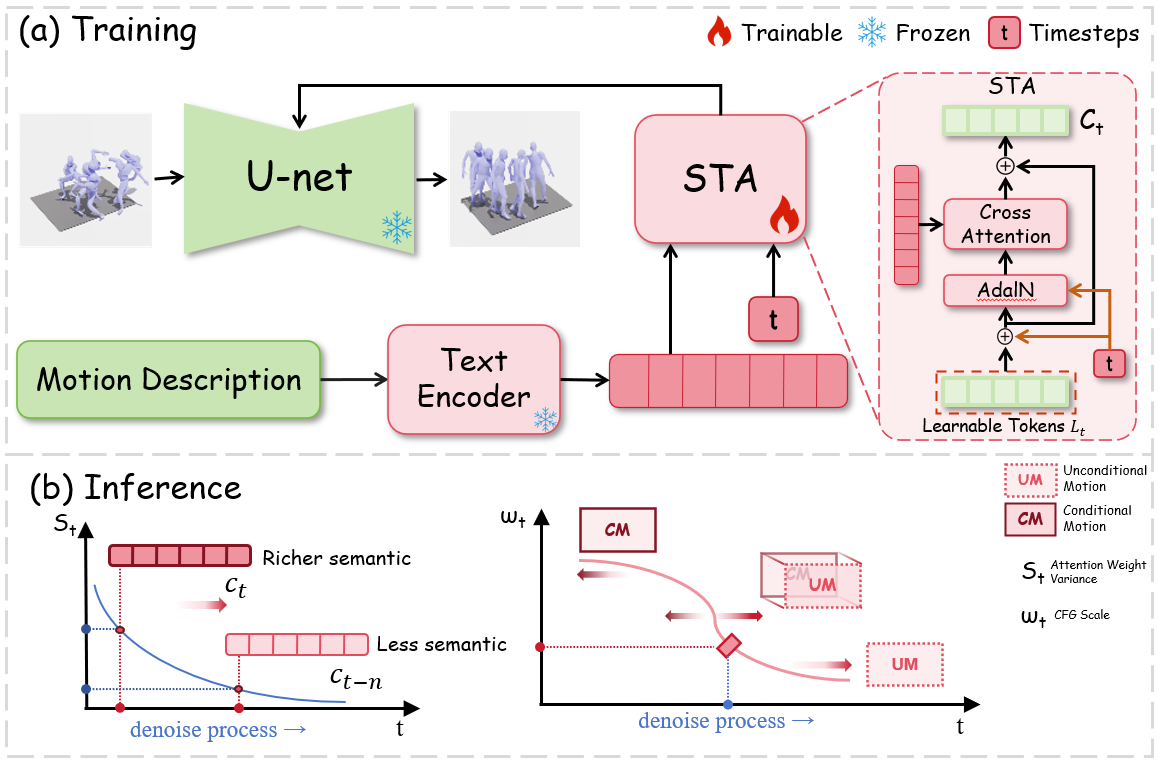}
    \vspace{-10pt}
    \caption{Overall Architecture of ANT. In part (a) Training, we introduce the Semantic Temporal Awareness (STA) module. STA is inserted between the text encoder and the U-Net, dynamically modulating the text features by attending to the diffusion timestep \( t \). In part (b) Inference, we observe that STA enables the model's attention to textual semantics to gradually decrease as sampling progresses. Based on this, we propose a CFG process that aligns with dynamic semantic adjustment: the CFG scale is progressively reduced during sampling, and a more efficient unconditional generation is applied in the later stages.}
    \label{fig:framework}
    \vspace{-0.5cm}
\end{figure*}

\section{Method}

\subsection{Preliminaries}
\subsubsection{\textbf{Text-to-Motion Process}}
We follow the diffusion framework of StableMoFusion \cite{huang2024stablemofusionrobustefficientdiffusionbased} for text-conditioned human motion generation. Let $c \in \mathbb{R}^{d_\text{c}}$ be a textual description encoded by a pretrained language model, where $d_\text{c}$ denotes the text embedding dimension. The target motion sequence $\mathbf{x}_0 \in \mathbb{R}^{N \times d_\text{m}}$ consists of $N$ frames, where each frame contains $d_\text{m}$-dimensional motion parameters.
The forward diffusion process progressively adds Gaussian noise to $\mathbf{x}_0$ over $T$ timesteps:
\begin{equation}
    \mathbf{x}_t = \sqrt{\alpha_t}\mathbf{x}_0 + \sqrt{1-\alpha_t}\boldsymbol{\epsilon}, \quad \boldsymbol{\epsilon} \sim \mathcal{N}(\mathbf{0}, \mathbf{I}),
\end{equation}
where $\{\alpha_t\}_{t=1}^T \in (0,1)^T$ is the noise schedule with $\alpha_t = \prod_{s=1}^t (1-\beta_s)$ and $\beta_s$ as the variance schedule.

In the reverse process, a motion-denoising network $G_\theta$ parameterized by $\theta$ is trained to predict the original motion $\mathbf{x}_0$ from the noisy input $\mathbf{x}_t$, conditioned on the timestep $t$ and text embedding $\mathbf{c}$:
\begin{equation}
    \min_\theta \mathbb{E}_{t,\mathbf{x}_0,\boldsymbol{\epsilon},\mathbf{c}} \bigl[ \|G_\theta(\mathbf{x}_t, t, \mathbf{c}) - \mathbf{x}_0\|_2^2 \bigr].
\end{equation}

During inference, $\mathbf{x}_0$ is generated by iteratively denoising from $\mathbf{x}_T \sim \mathcal{N}(0, \mathbf{I})$ using a sampler such as DPM-Solver++ \cite{lu2023dpmsolverfastsolverguided}, which accelerates generation by reducing the number of reverse steps from $T = 1000$ to as few as 10 steps. Given a textual prompt $\mathcal{T}$, the embedded text vector $\mathbf{c} = \text{CLIP}(\mathcal{T})$ is cached to avoid redundant computations. Classifier-Free Guidance (CFG) is applied to trade off text-motion alignment and motion fidelity:
\small{
\begin{equation}
    G_s(\mathbf{x}_t, t, \mathbf{c}) = G_\theta(\mathbf{x}_t, t, \emptyset) + \omega \cdot \bigl(G_\theta(\mathbf{x}_t, t, \mathbf{c}) - G_\theta(\mathbf{x}_t, t, \emptyset)\bigr),
\end{equation}}

\noindent where $\omega$ is the guidance scale and $G_\theta(\mathbf{x}_t, t, \emptyset)$ denotes unconditional prediction. Half-precision floating-point computation (FP16) further accelerates inference without compromising quality. This pipeline ensures efficient and high-quality generation of $N$-frame motion sequences within a fraction of the original computational cost.

\subsubsection{\textbf{The Denoising Process}}
In diffusion models, low-frequency components are recovered earlier than high-frequency components during the reverse denoising process \cite{boostingdiffusionmodelsmoving}. Formally, given a motion signal in the frequency domain $ \hat{m}_t(\omega) $, the signal-to-noise ratio (SNR) at frequency $ \omega $ is defined as:
\begin{equation}
    \text{SNR}(\omega) = \frac{|\hat{m}_0(\omega)|^2}{\int_0^t g^2(s) \, ds},
\end{equation}

\noindent where $ \hat{m}_0(\omega) $ is the initial power spectral density of the motion signal and $ \int_0^t g^2(s) \, ds $ represents the accumulated noise energy. For higher frequencies $ \omega_H $, the SNR decreases more rapidly than for lower frequencies $ \omega_L $, i.e., 
\begin{equation}
    \text{SNR}(\omega_H) < \text{SNR}(\omega_L), \quad \forall \, \omega_H > \omega_L.
\end{equation}

As a result, low-frequency components are restored first, providing a semantic foundation for the subsequent recovery of high-frequency details.

\subsubsection{\textbf{Low-Frequency Structure Dependence on High-Frequency Motion}}
\label{sec:theory 2}

In the reverse denoising process of diffusion models for motion generation, the accurate restoration of high-frequency motion components is structurally dependent on the consistent reconstruction of low-frequency components.

Let $\mathbf{m}_t$ denote the motion signal representation at timestep $t$. Let its frequency domain representation be partitioned into low-frequency components $\hat{\mathbf{m}}_{t,L}$ (corresponding to frequencies $\omega \in \Omega_L$) and high-frequency components $\hat{\mathbf{m}}_{t,H}$ (corresponding to frequencies $\omega \in \Omega_H$, where typically $\min(\Omega_H) > \max(\Omega_L)$). The reverse diffusion step estimates the posterior distribution $p(\mathbf{m}_{t-1} | \mathbf{m}_t)$.

The dependency implies that the uncertainty regarding the high-frequency components $\hat{\mathbf{m}}_{t-1, H}$ at step $t-1$, given the noisy observation $\mathbf{m}_t$, is reduced when conditioned on the concurrently estimated low-frequency components $\hat{\mathbf{m}}_{t-1, L}$. Formally, this relationship can be expressed via conditional variance reduction:
\begin{equation}
    \mathbb{E}_{\hat{\mathbf{m}}_{t-1, L} \sim p(\cdot | \mathbf{m}_t)} \left[ \text{Var}[\hat{\mathbf{m}}_{t-1, H} \mid \mathbf{m}_t, \hat{\mathbf{m}}_{t-1, L}] \right] < \text{Var}[\hat{\mathbf{m}}_{t-1, H} \mid \mathbf{m}_t].
    \label{eq:variance_reduction}
\end{equation}

This inequality holds assuming that $\hat{\mathbf{m}}_{t-1, H}$ and $\hat{\mathbf{m}}_{t-1, L}$ are not conditionally independent given $\mathbf{m}_t$. This dependency arises because the denoising network, in estimating $p(\mathbf{m}_{t-1} | \mathbf{m}_t)$, implicitly leverages the structural information inferred from the more robust low-frequency content within $\mathbf{m}_t$ (and consequently $\hat{\mathbf{m}}_{t-1, L}$) to guide the reconstruction of the high-frequency details $\hat{\mathbf{m}}_{t-1, H}$.

Furthermore, the inherent characteristics of signal corruption during the diffusion process amplify this dependency. The signal-to-noise ratio (SNR) typically decreases with increasing frequency, particularly in later diffusion steps (smaller $t$):
\begin{equation}
    \text{SNR}(\omega_H) \ll \text{SNR}(\omega_L), \quad \text{for typical } \omega_H \in \Omega_H, \omega_L \in \Omega_L.
    \label{eq:snr_inequality}
\end{equation}

Consequently, the high-frequency components within the noisy signal $\mathbf{m}_t$ are significantly more obscured by noise compared to the low-frequency components. Accurate restoration of $\hat{\mathbf{m}}_{t-1, H}$ therefore relies substantially on the contextual foundation provided by the simultaneously restored low-frequency structure $\hat{\mathbf{m}}_{t-1, L}$. This hierarchical reliance is crucial for generating motions that exhibit both semantic consistency (e.g., fine-grained gestures aligning with overall body posture and action) and spatial coherence (e.g., detailed limb movements respecting the constraints imposed by the larger skeletal configuration). 

\subsection{ANT Architecture}
The ANT architecture (Figure \ref{fig:framework}) optimizes the denoising steps for both semantic information and Classifier-Free Guidance (CFG). To fully explore the distinct roles of textual semantics in the denoising process of diffusion models, we introduce a Semantic Temporal Awareness (STA) module in Section \ref{sec:STA}. This module dynamically adjusts the text embeddings by incorporating timestep features, emphasizing low-frequency semantics in the early denoising stages and focusing on high-frequency details in the later stages. In Section \ref{sec:cfg}, based on observations of the attention to textual features in the ANT architecture, we design a dynamic planning strategy for Classifier-Free Guidance (CFG) that gradually weakens textual guidance during the denoising process.


\begin{figure}
    \centering
    \includegraphics[width=\linewidth]{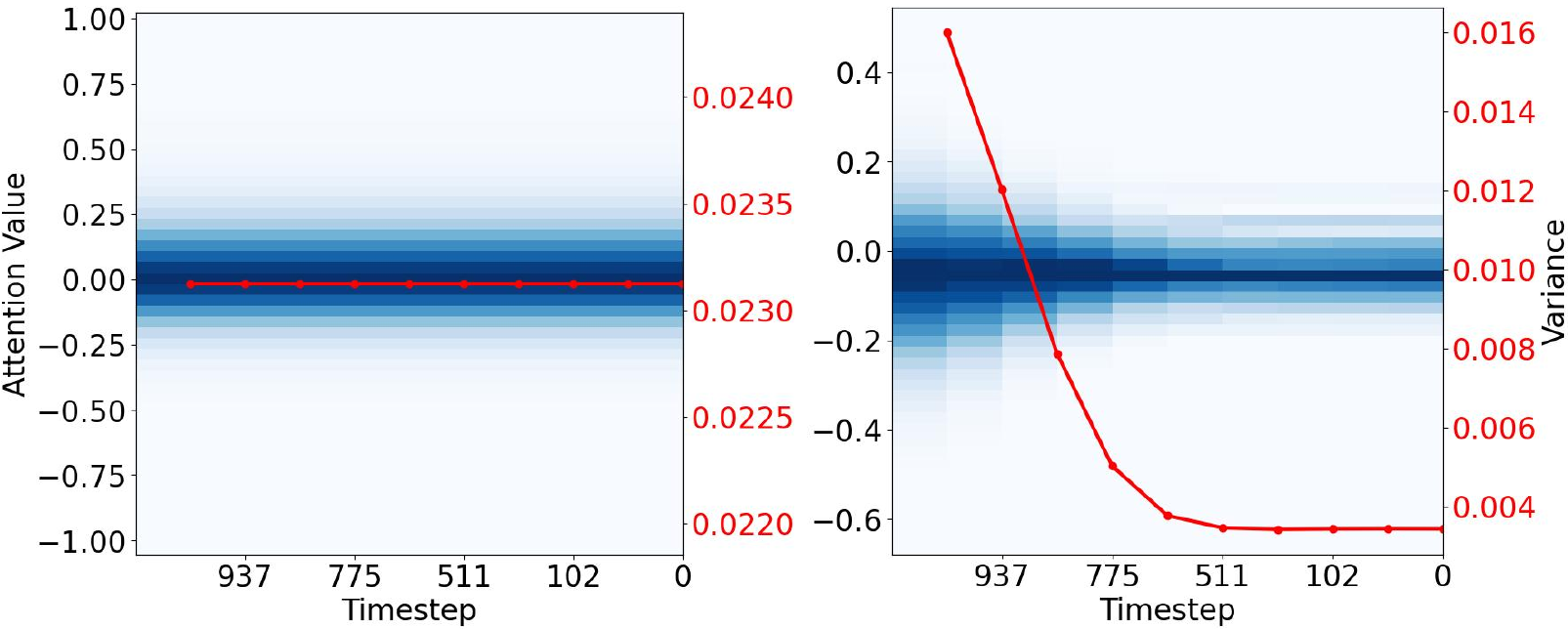}
    
    \caption{Distribution of attention weights in the UNet cross-attention module over the entire test set. The left panel corresponds to the baseline, and the right panel represents our method. In each panel, the red line denotes the variance of the attention weights, while the blue heatmap illustrates their distribution, with darker colors indicating a higher frequency of occurrence. The heatmap values have been normalized to a range of 0 to 1.}
    \label{fig:attention}
    
\end{figure}

\subsubsection{\textbf{Semantic Temporal Awareness Module}}
\label{sec:STA}
Unlike previous approaches \cite{huang2024stablemofusionrobustefficientdiffusionbased,mdm,guo2023momask,t2mgpt}, which directly use the text features \( c \) from a text encoder as conditions for predicting the output, we introduce the STA module. Positioned between the text features and the U-Net, this module is inspired by \cite{hu2024ella} and aims to design a connector that enhances the informativeness of the conditions used for noise prediction.  Below is the specific process of the STA:
The Learnable Tokens \( L \) are combined with the text features \( \mathbf{c} \) from the encoder to form a timestep-modulated feature \( L_t \):
\begin{equation}
L_t = L \oplus z_t.
\end{equation}

Here, \( z_t \) is the time feature and $\oplus$ denotes residual connection. This timestep-modulated feature is then processed through the Adaptive Layer Normalization (AdaLN) \cite{adaln}, which is defined by the following equation:
\begin{equation}
\hat{L}_t = \gamma \cdot \frac{L_t + \alpha(z_t)}{\sigma + \varepsilon} + \beta,
\end{equation}
where \( \gamma \), \( \alpha(t) \), \( \sigma \), and \( \beta \) are learnable parameters, and \( \varepsilon \) is a small constant to prevent division by zero. The output, after being processed through CrossAttention and a residual connection, is integrated with temporal information to produce the semantic feature with time-dependent text feature $\mathbf{c}_t$:
\begin{equation}
    \hat{c}_t = \hat{L}_t \oplus\text{CrossAttention}(c, \hat{L}_t).
\end{equation}

This process allows the model to dynamically incorporate both the temporal and semantic aspects, and unsupervisedly leverage the denoising prior to learn the semantic features at different time steps.



\subsubsection{\textbf{Adaptive Guidance Schedule}}
\label{sec:cfg}
Classifier-Free Guidance (CFG) is a widely used technique in generative models (such as diffusion models) that enhances conditional generation by computing both conditional and unconditional outputs and interpolating between them using a guidance scale. While prior work \cite{huang2024stablemofusionrobustefficientdiffusionbased,mdm} typically adopts a fixed guidance scale (e.g., $\omega = 4.5$) to improve semantic alignment, our denoising observations reveal that a static scale fails to adapt to the varying needs of different stages in the generation process.

As illustrated in the right of Figure \ref{fig:attention}, we visualize the distribution of attention values to text features in the ANT U-Net across the entire test set (in blue), and additionally report the variance of the attention values (in red). We observe that, as denoising progresses, the attention values increasingly concentrate around 0. This indicates that in the early stages, the network tends to focus more on capturing semantic information—reflected in a broader and more dispersed distribution of attention values across different tokens—while in the later stages, semantic attention diminishes, and the model gradually shifts toward unconditional generation. These insights suggest that stronger guidance should be applied during the early denoising steps, followed by a progressive reduction in guidance intensity, ultimately favoring unconditioned generation as the process converges.

Moreover, CFG increases computational overhead by requiring both conditional and unconditional generations. From an efficiency perspective, we further analyze that when the number of denoising steps is sufficiently large, the model has already captured sufficient semantic information. At this stage, the primary objective shifts to refining fine-grained, high-frequency details. Compared to conditional generation, unconditional generation is better at capturing the natural and fluid distribution of motion. Furthermore, based on Theorem \ref{sec:theory 2}, high-frequency information generation relies heavily on low-frequency components. This insight motivates us to forgo CFG in the later stages of denoising and instead perform more efficient unconditional generation. 


\begin{table*}[t]
\centering
\resizebox{\textwidth}{!}{%
\begin{tabular}{llccccccc}
\toprule
\textbf{Method} & \textbf{Venue} & \textbf{FID $\downarrow$}  &\multicolumn{3}{c}{\textbf{R-Precision $\uparrow$}} & \textbf{Diversity $\rightarrow$}  & \textbf{MM-Dist$\downarrow$} & \textbf{Multimodality$\uparrow$}\\ \cmidrule(lr){4-6}
 &  &  & \textbf{Top1} & \textbf{Top2} & \textbf{Top3} & & & \\ \midrule
\multicolumn{9}{c}{\textbf{HumanML3D}} \\ \midrule
Ground Truth   &  - & $0.002^{\pm0.000}$ & $0.511^{\pm0.003}$ & $0.703^{\pm0.003}$ & $0.797^{\pm0.002}$ & $9.503^{\pm0.065}$ & - & -\\ 
MLD \cite{chen2023executingcommandsmotiondiffusion} & CVPR 2023   & $0.473^{\pm0.013}$ & $0.481^{\pm0.003}$ & $0.673^{\pm0.003}$ & $0.772^{\pm0.002}$ & $9.724^{\pm0.082}$& $3.196^{\pm0.010}$ & $2.413^{\pm0.079}$\\
ReMoDiffuse \cite{zhang2023remodiffuse} & ICCV 2023 & $0.103^{\pm0.004}$ & $0.510^{\pm0.005}$ & $0.698^{\pm0.006}$ & $0.795^{\pm0.004}$ & $9.018^{\pm0.075}$ & $3.025^{\pm0.008}$ & $1.795^{\pm0.043}$\\
MotionDiffuse \cite{zhang2024motiondiffuse} & TPAMI 2024 & $0.630^{\pm0.001}$ & $0.491^{\pm0.001}$ & $0.681^{\pm0.001}$ & $0.782^{\pm0.001}$ & $9.410^{\pm0.049}$ & $3.113^{\pm0.001}$ & $1.553^{\pm0.042}$\\
MotionLCM \cite{dai2024motionlcmrealtimecontrollablemotion} & ECCV 2024 & $0.467^{\pm0.012}$ & $0.502^{\pm0.003}$ & $0.701^{\pm0.002}$ & $0.803^{\pm0.002}$ & $9.361^{\pm0.660}$ &$3.012^{\pm0.007}$ &$2.172^{\pm0.082}$\\
T2M-GPT \cite{t2mgpt} & CVPR 2023 & $0.141^{\pm0.005}$ & $0.492^{\pm0.003}$ & $0.679^{\pm0.002}$ & $0.775^{\pm0.002}$ & $9.722^{\pm0.082}$ & $3.121^{\pm0.009}$ &$1.831^{\pm0.048}$\\
MMM \cite{pinyoanuntapong2024mmmgenerativemaskedmotion}& CVPR 2024 & $0.089^{\pm0.002}$ & $0.515^{\pm0.002}$ & $0.708^{\pm0.002}$ & $0.804^{\pm0.002}$& $9.577^{\pm0.050}$ &$2.926^{\pm0.007}$ &$1.226^{\pm0.035}$\\
MoMask \cite{guo2023momask}& CVPR 2024 & \textcolor{red}{$0.045^{\pm0.002}$} & $0.521^{\pm0.002}$ & $0.713^{\pm0.002}$ & $0.807^{\pm0.002}$& - & $2.958^{\pm0.008}$ & $1.241^{\pm0.040}$\\
MDM \cite{mdm} & ICLR 2023   & $0.544^{\pm0.044}$ & $0.320^{\pm0.005}$ & $0.498^{\pm0.004}$ & $0.611^{\pm0.007}$ & \textcolor{blue}{$9.559^{\pm0.086}$} &$5.556^{\pm0.027}$ &\textcolor{red}{$2.799^{\pm0.072}$} \\
StableMoFusion \cite{huang2024stablemofusionrobustefficientdiffusionbased} & ACM MM 2024 & $0.152^{\pm0.004}$ & $0.546^{\pm0.002}$ & $0.742^{\pm0.002}$ & $0.835^{\pm0.002}$ & \textcolor{red}{$9.466^{\pm0.002}$} & \textcolor{blue}{$2.781^{\pm0.011}$} &$1.362^{\pm0.062}$\\ 
StableMoFusion Efficiency & ACM MM 2024 & $2.845^{\pm0.027}$ & $0.401^{\pm0.003}$ & $0.599^{\pm0.003}$ & $0.719^{\pm0.003}$ & $8.699^{\pm0.098}$ & - &$2.276^{\pm0.065}$ \\ \midrule
ANT (Ours, on MDM) & - & $0.377^{\pm0.038}$ & $0.456^{\pm 0.006}$ & $0.656^{\pm 0.007}$ & $0.763^{\pm0.006}$ & $9.886^{\pm0.068}$ & - &  \textcolor{blue}{$2.595^{\pm0.006}$}\\
ANT (Ours, on StableMoFusion) & - & $0.099^{\pm0.004}$ & \textcolor{blue}{$0.560^{\pm0.002}$} & \textcolor{blue}{$0.751^{\pm0.003}$} & \textcolor{blue}{$0.841^{\pm0.002}$} & $9.585^{\pm0.090}$ & $2.875^{\pm0.015}$ & $1.874^{\pm0.062}$ \\
ANT (Ours, on StableMoFusion, w/o DCFG) & - & \textcolor{blue}{$0.071^{\pm0.004}$} & \textcolor{red}{$0.565^{\pm0.006}$} & \textcolor{red}{$0.756^{\pm0.003}$} & \textcolor{red}{$0.843^{\pm0.002}$} & $9.585^{\pm0.090}$ & \textcolor{red}{$2.763^{\pm0.016}$} & $1.827^{\pm0.110}$\\ 
\midrule
\multicolumn{9}{c}{\textbf{KIT-ML}} \\ \midrule
Ground Truth & - & $0.031^{\pm0.004}$ & $0.424^{\pm0.005}$ & $0.649^{\pm0.006}$ & $0.779^{\pm0.006}$ & $11.080^{\pm0.097}$ & - & - \\ 
MLD \cite{chen2023executingcommandsmotiondiffusion} & CVPR 2023 & $0.404^{\pm0.027}$ & $0.390^{\pm0.008}$ & $0.609^{\pm0.008}$ & $0.734^{\pm0.007}$ & $10.800^{\pm0.117}$ & $3.204^{\pm0.027}$ & \textcolor{blue}{$2.192^{\pm0.071}$}\\
ReMoDiffuse \cite{zhang2023remodiffuse} & ICCV 2023 & $0.155^{\pm0.006}$ & $0.427^{\pm0.014}$ & $0.641^{\pm0.004}$ & $0.765^{\pm0.055}$ & $10.800^{\pm0.105}$ & \textcolor{red}{$1.239^{\pm0.028}$} & $1.239^{\pm0.028}$ \\
MotionDiffuse \cite{zhang2024motiondiffuse} & TPAMI 2024  & $1.954^{\pm0.062}$ & $0.417^{\pm0.004}$ & $0.621^{\pm0.004}$ & $0.739^{\pm0.004}$ & \textcolor{red}{$11.100^{\pm0.143}$} & $2.958^{\pm0.005}$ & $0.730^{\pm0.013}$ \\
T2M-GPT \cite{t2mgpt} & CVPR 2023 & $0.514^{\pm0.029}$ & $0.416^{\pm0.006}$ & $0.627^{\pm0.006}$ & $0.745^{\pm0.006}$ & $10.921^{\pm0.108}$ & $3.007^{\pm0.023}$ & $1.570^{\pm0.039}$ \\
MotionGPT \cite{jiang2023motiongpthumanmotionforeign} & NeurIPS 2023 & $0.510^{\pm0.016}$ & $0.366^{\pm0.005}$ & $0.558^{\pm0.004}$ & $0.680^{\pm0.005}$ & $10.350^{\pm0.084}$ & $3.527^{\pm0.021}$ & \textcolor{red}{$2.328^{\pm0.117}$} \\
MMM \cite{pinyoanuntapong2024mmmgenerativemaskedmotion}& CVPR 2024& $0.316^{\pm0.028}$ & $0.404^{\pm0.005}$ & $0.621^{\pm0.005}$ & $0.744^{\pm0.004}$ & $10.910^{\pm0.101}$ & $2.977^{\pm0.019}$ & $1.232^{\pm0.039}$ \\
MoMask \cite{guo2023momask}& CVPR 2024 & \textcolor{red}{$0.204^{\pm0.011}$} & $0.433^{\pm0.007}$ & $0.656^{\pm0.005}$ & $0.781^{\pm0.005}$ & - & $2.779^{\pm0.022}$ & $1.131^{\pm0.043}$ \\
MDM \cite{mdm} & ICLR 2023 & $0.497^{\pm0.021}$ & $0.164^{\pm0.004}$ & $0.291^{\pm0.004}$ & $0.396^{\pm0.004}$ & $10.847^{\pm0.119}$ & $9.191^{\pm0.022}$ & $1.907^{\pm0.214}$ \\
StableMoFusion \cite{huang2024stablemofusionrobustefficientdiffusionbased} & ACM MM 2024 & $0.258^{\pm0.029}$ & \textcolor{blue}{$0.445^{\pm0.006}$} & \textcolor{blue}{$0.660^{\pm0.005}$} & \textcolor{blue}{$0.782^{\pm0.004}$} & $10.936^{\pm0.077}$ & $2.800^{\pm0.018}$ & $1.362^{\pm0.062}$ \\ \midrule
ANT (Ours) & - & \textcolor{blue}{$0.236^{\pm0.015}$} & \textcolor{red}{$0.465^{\pm0.007}$} & \textcolor{red}{$0.694^{\pm0.006}$} & \textcolor{red}{$0.813^{\pm0.005}$} & \textcolor{blue}{$11.029^{\pm0.102}$} & \textcolor{blue}{$2.689^{\pm0.020}$} & $1.578^{\pm0.056}$ \\
\bottomrule
\end{tabular}%
}
\caption{Quantitative comparison on the HumanML3D and KIT-ML datasets. $\pm$ indicates a 95\% confidence interval. $\downarrow$: Lower is better. $\uparrow$: Higher is better. $\rightarrow$: Closer to the Ground Truth (GT) is better. \textcolor{red}{Red} and \textcolor{blue}{Blue} indicate the best and the second-best results respectively across all methods for each metric. Our method, ANT, demonstrates state-of-the-art or highly competitive performance across multiple key metrics on both datasets.}
\label{tab:combined_results}
\end{table*}

To implement this adaptive guidance mechanism, we define a time-dependent guidance scale \(\omega_t\) as follows:

\begin{equation}
    \omega_t = \omega_{\min} + \phi(t)(\omega_{\max} - \omega_{\min}),
\end{equation}

\noindent where \(\omega_{\max}\) and \(\omega_{\min}\) denote the maximum and minimum guidance scales respectively, and \(\phi(t)\) is a monotonically decreasing function that controls the decay of guidance strength over the denoising timestep \(t\). In this paper, We leverage cosine schedule as $\phi(t)$. 

\begin{equation}
   \omega _t = max\{\omega _{\min } + \frac {1}{2}\left ( 1 + \cos (\lambda\frac {T-t}{T}\pi )\right )(\omega _{\max } - \omega _{\min }) , 0\}.
\end{equation}

Here, $\lambda$ denotes the period coefficient, and $T$ represents the total number of time steps. Accordingly, the noise prediction at timestep \(t\) with Classifier-Free Guidance is computed as:

\begin{equation}
    \hat{\epsilon}_{\text{CFG}} = \hat{\epsilon}_{\text{uncond}} + \omega_t \cdot \left(\hat{\epsilon}_{\text{cond}} - \hat{\epsilon}_{\text{uncond}}\right),
\end{equation}

\noindent where \(\hat{\epsilon}_{\text{cond}}\) and \(\hat{\epsilon}_{\text{uncond}}\) represent the conditional and unconditional denoising predictions, respectively.

To further improve efficiency during the late stages of denoising where semantic conditioning becomes less influential, we omit the conditional branch altogether when \(t\) exceeds a certain threshold (e.g., \(t > 0.5T\)). In such cases, we use:

\begin{equation}
    \hat{\epsilon}_t = \hat{\epsilon}_{\text{uncond}}.
\end{equation}

Thereby reducing computational overhead while promoting the generation of coherent high-frequency details aligned with the learned data distribution.

\begin{figure*}[t]
    \centering
    \vspace{-5pt}
    \includegraphics[width=\textwidth]{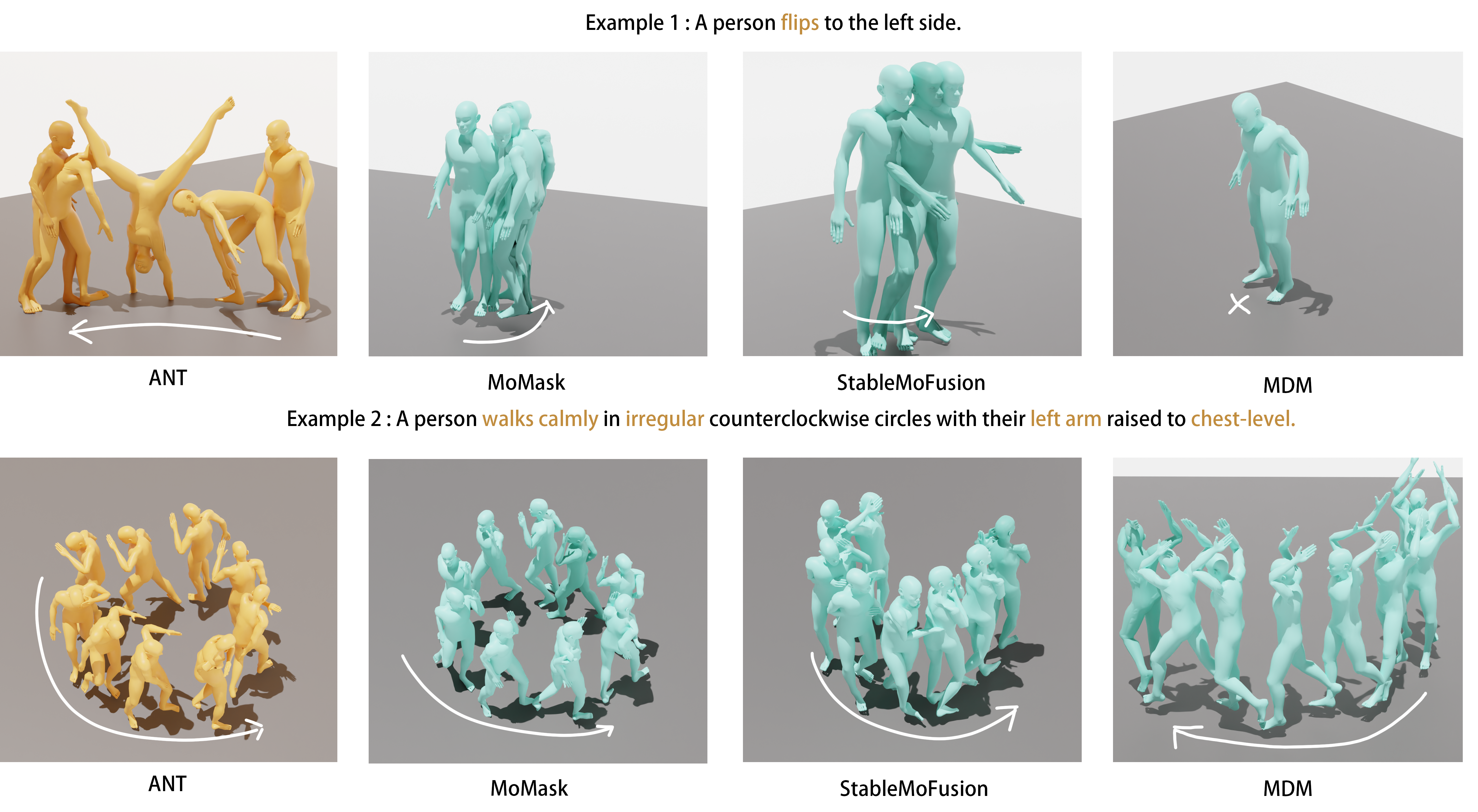}
    \vspace{-10pt}
    \caption{Visualization Comparison. We compare the visual results of ANT with other three state-of-the-art methods. In both examples, ANT consistently demonstrates more \uline{accurate}, \uline{natural}, and \uline{fine-grained} motion generation compared to the others.}
    \label{fig:visualization}
\end{figure*}

\section{Experiments}

We evaluate our approach on two standard motion-language
benchmarks: HumanML3D \cite{Guo_2022_CVPR} and KIT-ML \cite{Plappert2016}. HumanML3D con-
tains 14,616 motion sequences from AMASS \cite{AMASS:ICCV:2019}  and HumanAct12 \cite{guo2020action2motion}, each paired
with three text descriptions (44,970 total), covering diverse
actions like walking, exercising, and dancing. KIT-ML in-
cludes 3,911 motions and 6,278 descriptions, serving as a
smaller-scale benchmark. We follow StableMoFusion’s pose
representation and apply mirroring-based augmentation. Data
is split into training, validation, and test sets with a ratio of
0.8:0.15:0.05.
\noindent \textbf{Evaluation Metrics.}
In addition to the commonly utilized metrics such as Frechet Inception Distance (FID) \cite{FID}, R-Precision, Multimodal Distance (MM-Dist), and Diversity, which are employed by StableMoFusion \cite{huang2024stablemofusionrobustefficientdiffusionbased}. Furthermore, human evaluation is employed to obtain accuracy and human preference results for the outputs generated by the model.


    
    
\subsection{Experimental Setup}
We adopt model architecture settings similar to those of MDM \cite{mdm} and StableMoFusion \cite{huang2024stablemofusionrobustefficientdiffusionbased}. For MDM, we use a batch size of 64 and the AdamW \cite{loshchilov2019decoupledweightdecayregularization} optimizer. Our models are trained with a time step of \( T = 50 \), following a cosine noise schedule. The total number of training iterations is fixed at 120,000, with a learning rate of \( 1 \times 10^{-4} \).  For StableMoFusion, we adhere to its training methodology, running for 200,000 steps. The time step is set to \( T = 50 \), and we employ DPM-Solver for inference, using 10 actual sampling steps. The $\lambda$ is set to 1.5 and unconditional generation start time is set to 0.5T. $\omega_{max}$ and $\omega_{min}$ is set to 3.0 and 1.5, The optimal values for $\omega_{max}$ and $\omega_{min}$ are selected via grid search on the validation set, selecting those that achieved the best Top-1 performance (See detail in the appendix F),  respectively. The entire training process can be efficiently executed on a single RTX 4090 GPU with 24 GB of memory.

\begin{table}[t]
\centering
\resizebox{\columnwidth}{!}{%
\begin{tabular}{lcccc}
\hline
\textbf{Method}  & \textbf{Top1} & \textbf{Top2} & \textbf{Top3} \\ \hline
BERT (w/o STA)        & $0.547^{\pm 0.003}$  & $0.742^{\pm0.002}$  & $0.835^{\pm0.002}$    \\ 
BERT (w STA)       & $0.551^{\pm0.002}$  & $0.747^{\pm0.003}$  & $0.836^{\pm0.002}$    \\[0.5em]
MoCLIP (w/o STA)      & $0.528^{\pm0.002}$  & $0.718^{\pm0.002}$  & $0.812^{\pm0.002}$    \\ 
MoCLIP (w STA)     & $0.554^{\pm0.003}$  & $0.747^{\pm0.003}$  & $0.838^{\pm0.002}$    \\[0.5em]
LongCLIP (w/o STA)    & $0.528^{\pm0.002}$  & $0.718^{\pm0.002}$  & $0.812^{\pm0.002}$    \\ 
LongCLIP (w STA)   & $0.513^{\pm0.003}$  & $0.702^{\pm0.002}$  & $0.797^{\pm0.003}$    \\[0.5em]
CLIP (w/o STA)        & $0.538^{\pm0.003}$  & $0.730^{\pm0.002}$  & $0.822^{\pm0.002}$    \\ 
CLIP (w STA)       & $0.523^{\pm0.003}$  & $0.717^{\pm0.002}$  & $0.812^{\pm0.002}$    \\[0.5em]
T5 (w/o STA)          & $0.549^{\pm0.004}$  & $0.747^{\pm0.004}$  & $0.838^{\pm0.003}$    \\ 
\rowcolor{gray!20}
T5 (w STA)         & $0.565^{\pm0.006}$  & $0.756^{\pm0.003}$  & $0.843^{\pm0.002}$    \\ \hline
\end{tabular}%
}
\caption{Experimental results for different methods on text-to-motion generation. The term "w STA" indicates the use of an STA. All values are reported as mean and $\pm$ indicates a 95\% confidence interval.}
\label{table:text_encoder_ablation}
\vspace{-1.0cm}
\end{table}

\subsection{Comparison to State-of-the-art Approaches}

\textbf{Quantitative comparisons.} \label{sec:com_sota}
Following \cite{t2mgpt,guo2023momask}, we report the average over 20 repeated generations with a 95\% confidence interval. Table \ref{tab:combined_results} presents evaluations on the HumanML3D \cite{Guo_2022_CVPR} and KIT-ML \cite{Plappert2016} datasets, respectively, in comparison with state-of-the-art (SOTA) approaches.

In terms of improvement over the baseline, ANT demonstrates strong performance, achieving substantial gains on metrics such as FID (HumanML3D: \textbf{0.071} vs. 0.152; KIT-ML: \textbf{0.236} vs. 0.258) and R-Precision (Top-1 HumanML3D: \textbf{0.565} vs. 0.546; KIT-ML: \textbf{0.465} vs. 0.445). This indicates that ANT can significantly enhance the performance of the baseline model. When compared with SOTA VQ-based models such as MMM and MoMask, our method still yields competitive results on HumanML3D and KIT-ML. Although our FID is slightly higher than that of MoMask, our R-Precision surpasses all existing SOTA models. It is worth noting that the baseline model we used performs significantly worse than SOTA methods like MMM. However, with our proposed ANT enhancements, we achieve competitive results. \\
\begin{table}[t]
\centering
\begin{tabular}{lcc}
\toprule
\textbf{Method} & \textbf{Average Time (s)}  \\[0.5ex] 
\midrule
ANT (w/o DCFG)    & 0.949  \\ 
Baseline method (w/o DCFG)    & 0.878  \\ 
ANT (w DCFG) & 0.741   \\ 
Baseline method (w DCFG) & 0.717   \\ 
\bottomrule
\end{tabular}
\caption{Average Processing Times per 32-Batch (use T5 text encoder).} 
\label{tab:processing_times}
\vspace{-0.7cm}
\end{table}

\textbf{Qualitative comparison.} Figure \ref{fig:visualization} shows the visual comparison between our model and other state-of-the-art methods under the same prompts. For the prompt "A person flips to the left side," MDM, StableMoFusion, and MoMask all produce incorrect motions, indicating a lack of semantic alignment. In contrast, our model generates natural and smooth motion that accurately reflects the input text. For the more detailed prompt, "A person walks calmly in irregular counterclockwise with their left arm raised to chest-level", StableMoFusion and MDM incorrectly raises the arm, while MoMask not only overlooks the meaning of "irregular" but also raises the left arm higher than chest level. Our model successfully captures all the fine-grained information and produces motion that remains faithful to the prompt. Overall, the visualization results demonstrate that our approach outperforms baselines and other models in terms of semantic accuracy, motion fluency, and attention to detail. \\

\textbf{ANT Boosts Performance on Complex Text Descriptions.} To evaluate our model's performance on semantically rich text, we select prompts from the HumanML3D test set that are longer than the average length and contain at least two verbs or include adverbs within a single sentence. Table \ref{table:hard_text} reports the results before and after applying ANT. As shown, ANT significantly improves all metrics on long and fine-grained prompts compared to the baseline. This demonstrates the strong advantages of our model in text comprehension, fine-grained motion generation, and handling long textual inputs. \\

\textbf{Plug-and-Play Architecture.} We validate the effectiveness of ANT not only on the StableMoFusion (DPM-Solver) but also on the MDM (DDIM). The results are shown in Table \ref{tab:combined_results}. ANT (MDM) achieves significant improvements over the baseline in both FID and R-Precision (FID: 0.377 vs. 0.544; Top-1: 0.456 vs. 0.320). These results confirm the adaptability of our method to different diffusion architectures and demonstrate its potential as a practical enhancement for diffusion-based text-to-motion models.\\

\begin{figure}
    \centering
    \includegraphics[width=\linewidth]{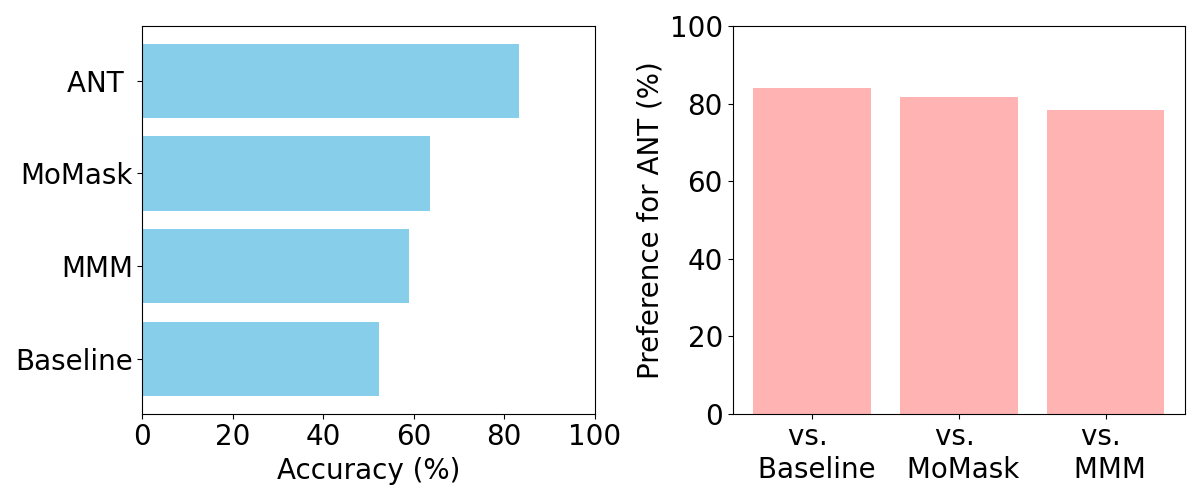}
    \vspace{-10pt}
    \caption{Comparative performance of ANT (StableMoFusion) versus other baseline methods based on human evaluation. The left panel depicts the accuracy achieved during manual assessments. The right panel illustrates the win rate when compared against other baseline methods. vs.Baseline means against original StableMoFusion.}
    \label{tab:human_eval}
    \vspace{-0.5cm}
\end{figure}


\textbf{Efficiency.} In table \ref{tab:processing_times}, we compare the efficiency of two our methods and baseline methods under the same text encoder and batch conditions. Our approach (ANT) shows a relatively significant performance improvement over the baseline, given the relatively small overhead of its implementation. 
Furthermore, based on insights gained from the ANT method, we applied unconditional generation at a later stage to improve efficiency, as indicated by the comparison between ANT and baseline method with efficiency sampling. 
Under conditions of limited performance degradation, this approach provides a significant enhancement, representing a favorable trade-off between efficiency and performance. 
In summary, the experimental results confirm that the ANT-based method is effective for achieving notable performance gains with minimal additional overhead, making it a viable solution for applications demanding both high performance and efficiency.

\textbf{Human Evaluation.}
\label{sec:human eval}
We randomly selected 100 text descriptions from the HumanML3D test set and invited 20 participants to subjectively evaluate the motion sequences generated by four models: ANT (StableMoFusion), StableMoFusion, MoMask, and MMM. The participants, unaware of the model names, were asked to rate whether the generated motions were semantically accurate, providing a "yes" or "no" response. We then calculated the average accuracy of motion generation for each model. Additionally, in the Pairwise Preference task, for each text description, we presented the results from ANT and one of the other three models (StableMoFusion, MoMask, or MMM), and asked the participants to choose the more natural and coherent motion from the two options. Each model pair was evaluated 100 times, resulting in a total of 300 binary comparisons. The Figure \ref{tab:human_eval} shows that ANT leads in motion accuracy, achieving 83.2\%. In terms of subjective preference, ANT also consistently outperformed the other models, securing an average preference rate of 81.3\%. ANT demonstrated significantly stronger semantic understanding and superior motion generation quality.

\subsection{Ablation Study}
\textbf{Analysis of Architectural Contributions.}
As shown in Table \ref{tab:combined_results}, we validate the effectiveness of STA. The improvement brought by STA alone over StableMoFusion has been discussed in Section \ref{sec:com_sota}. In Table \ref{tab:processing_times}, the sampling time for a single batch is reduced from 0.949s to 0.741s when using DCFG. This results in a 21.9\% increase in efficiency, with almost no loss in accuracy. This demonstrates the effectiveness of the DCFG when integrated into the STA architecture.
\begin{table}[t]
\centering
\resizebox{\columnwidth}{!}{%
\begin{tabular}{lccccc}
\hline
\textbf{Method}       & \textbf{FID}              & \textbf{Top1}              & \textbf{Top2}              & \textbf{Top3}              & \textbf{MultiModality} \\ \hline
GT                    & $0.015^{\pm0.008}$         & $0.4643^{\pm0.006}$        & $0.661^{\pm0.005}$         & $0.757^{\pm0.005}$         & -                   \\[0.5em]
StableMoFusion        & $1.316^{\pm0.040}$         & $0.394^{\pm0.002}$         & $0.576^{\pm0.004}$         & $0.680^{\pm0.003}$         & $1.873^{\pm0.052}$     \\[0.5em]
\rowcolor{gray!20}
ANT                   & $0.372^{\pm0.017}$         & $0.505^{\pm0.009}$         & $0.700^{\pm0.006}$         & $0.794^{\pm0.007}$         & $2.009^{\pm0.066}$     \\ \hline
\end{tabular}%
}
\caption{Experimental results for fine-grained generation of ANT. All values are reported as mean and $\pm$ indicates a 95\% confidence interval.}
\label{table:hard_text}
\vspace{-10pt}
\end{table}

\begin{table}[t]
\centering
\resizebox{\columnwidth}{!}{%
\begin{tabular}{lcccc}
\hline
\textbf{Method} & \textbf{FID} & \textbf{Top1} & \textbf{Top2} & \textbf{Top3} \\ \hline
Baseline    & $0.152^{\pm0.004}$      & $0.546^{\pm0.002}$      & $0.742^{\pm0.002}$      & $0.835^{\pm0.002}$      \\[0.5em]
Resampler   & $0.101^{\pm0.004}$      & $0.563^{\pm0.003}$      & \cellcolor{gray!20}{$\mathbf{0.756^{\pm0.003}}$}      & \cellcolor{gray!20}{$\mathbf{0.844^{\pm0.002}}$}      \\[0.5em]
Abstractor  & \cellcolor{gray!20}{$0.071^{\pm0.004}$}      & \cellcolor{gray!20}{$\mathbf{0.565^{\pm0.006}}$}      & \cellcolor{gray!20}{$\mathbf{0.756^{\pm0.003}}$}      & $0.843^{\pm0.002}$      \\ \hline
\end{tabular}%
}
\caption{Experimental results for different architecture of STA. All values are reported as mean and $\pm$ indicates a 95\% confidence interval.}
\label{table:sta_ablation}
\vspace{-0.8cm}
\end{table}
We also observe that applying DCFG directly to the baseline does not work. The baseline fails to distinguish between the early and late stages of the denoising process (Figure \ref{fig:attention}, due to its lack of temporal text awareness. This comparison further shows that the ANT architecture can effectively leverage the denoising prior in diffusion to improve prediction quality.\\
\textbf{Text Encoder.} In Table \ref{table:text_encoder_ablation}, we investigate the impact of applying ANT to various text encoders. We observe consistent performance improvements when using T5 \cite{raffel2023exploringlimitstransferlearning}, BERT \cite{devlin2019bertpretrainingdeepbidirectional}, and MoCLIP. In contrast, models based on CLIP and LongCLIP \cite{zhang2024longclipunlockinglongtextcapability} show performance degradation. This discrepancy can be attributed to the varying capacities of these models to capture fine-grained textual information. Large-scale encoders such as BERT and T5 benefit from rich pretraining, enabling them to generate detailed text representations. These representations facilitate dynamic modulation across time steps and support hierarchical, nuanced semantic understanding. On the other hand, CLIP and LongCLIP tend to produce coarser textual features, which often leads to semantic misalignment during dynamic processing. In contrast, MoCLIP demonstrates stronger alignment with motion semantics, as it has been fine-tuned on motion-specific datasets. This makes it more suitable for tasks that require precise and temporally coherent semantic modulation. Based on these findings, we adopt T5 as the text encoder in our work, as it delivers the strongest overall performance.\\
\textbf{STA Architecture.} 
In Table \ref{table:sta_ablation}, we explore two different architectures for the STA module. We conduct experiments on StableMoFusion and evaluate performance using FID and R-Precision. Both STA architectures demonstrate significant improvements over the baseline across all metrics while achieving comparable results in semantic alignment (R-Precision). Notably, Abstractor \cite{abstractor} outperforms Resampler \cite{resampler} in terms of FID (\textbf{0.071} vs. 0.101). Therefore, we adopt Abstractor in this work as the method for integrating semantic features and temporal steps.

\section{Conclusion}
In this paper, we design semantic temporal-aware methods for both the training (STA) and inference (DCFG) stages, based on the unique denoising mechanism of diffusion. Our method provides plug-and-play functionality, achieving more precise semantic alignment and more efficient sampling, demonstrating the potential of diffusion-based methods in T2M.

\clearpage
\bibliographystyle{ACM-Reference-Format}
\balance 
\bibliography{sample-base}

\clearpage
\appendix

\appendix
\section{Spectral Analysis of the Diffusion Process}
\label{sec:SpectralAnalysis}

We analyze the spectral properties of the signal during the diffusion process, focusing on how noise affects different frequency components over time.

\subsection{Power Spectral Density Evolution}

Consider the forward diffusion process described by the stochastic differential equation (SDE), simplified by assuming zero drift ($\mathbf{f}(\mathbf{m}_t, t) = 0$):
\begin{equation}
    \mathrm{d}\mathbf{m}_t = g(t)\mathrm{d}\mathbf{w}_t,
    \label{eq:sde_simple}
\end{equation}
where $g(t)$ is the diffusion coefficient and $\mathbf{w}_t$ is a standard Wiener process. The solution integrates to:
\begin{equation}
    \mathbf{m}_t = \mathbf{m}_0 + \boldsymbol{\epsilon}_t, \quad \text{where} \quad \boldsymbol{\epsilon}_t = \int_0^t g(s)\mathrm{d}\mathbf{w}_s.
    \label{eq:integral_form}
\end{equation}

Here, $\mathbf{m}_0$ is the initial clean signal and $\boldsymbol{\epsilon}_t$ represents the accumulated noise up to time $t$.

\begin{theorem}[Power Spectral Density in Simplified Diffusion]
\label{theorem:psd_simplified}
For the process defined by Eq. \eqref{eq:sde_simple}, the power spectral density (PSD) $S_{\mathbf{m}_t}(\omega)$ of the signal $\mathbf{m}_t$ at time $t$ is given by:
\begin{equation}
    S_{\mathbf{m}_t}(\omega) = |\hat{\mathbf{m}}_0(\omega)|^2 + \int_0^t g^2(s)\mathrm{d}s,
    \label{eq:psd_formula}
\end{equation}
where $\hat{\mathbf{m}}_0(\omega)$ is the Fourier transform of the initial signal $\mathbf{m}_0$.
\end{theorem}

\begin{proof}[Proof Sketch]
Applying the Fourier transform $\mathcal{F}$ to Eq. \eqref{eq:integral_form}, we get:
\begin{equation}
    \hat{\mathbf{m}}_t(\omega) = \mathcal{F}[\mathbf{m}_t](\omega) = \hat{\mathbf{m}}_0(\omega) + \hat{\boldsymbol{\epsilon}}_t(\omega).
\end{equation}

The PSD is defined as $S_{\mathbf{m}_t}(\omega) = \mathbb{E}[|\hat{\mathbf{m}}_t(\omega)|^2]$. Substituting the above:

\begin{align}
S_{\mathbf{m}_t}(\omega)
  &= \mathbb{E}\!\left[\left|\hat{\mathbf{m}}_0(\omega)+\hat{\boldsymbol{\epsilon}}_t(\omega)\right|^{2}\right] \nonumber\\
  &= \mathbb{E}\!\left[\left|\hat{\mathbf{m}}_0(\omega)\right|^{2}
     + 2\,\operatorname{Re}\!\bigl(\hat{\mathbf{m}}_0(\omega)\hat{\boldsymbol{\epsilon}}_t^{*}(\omega)\bigr)
     + \left|\hat{\boldsymbol{\epsilon}}_t(\omega)\right|^{2}\right].
\end{align}

Assuming the initial signal $\mathbf{m}_0$ is deterministic or independent of the subsequent noise $\boldsymbol{\epsilon}_t$, and noting that $\mathbb{E}[\hat{\boldsymbol{\epsilon}}_t(\omega)] = 0$, the cross-term vanishes: $\mathbb{E}[\hat{\mathbf{m}}_0(\omega)\hat{\boldsymbol{\epsilon}}_t^*(\omega)] = \hat{\mathbf{m}}_0(\omega)\mathbb{E}[\hat{\boldsymbol{\epsilon}}_t^*(\omega)] = 0$.
The expected energy of the noise in the frequency domain is a standard result from the properties of Itô integrals (related to the autocorrelation of the Wiener process):
\begin{equation}
    \mathbb{E}[|\hat{\boldsymbol{\epsilon}}_t(\omega)|^2] = \int_0^t g^2(s)\mathrm{d}s.
    \label{eq:noise_energy_freq}
\end{equation}

This noise energy term is independent of frequency $\omega$, characteristic of white noise accumulation in the frequency domain. Combining these results yields Eq. \eqref{eq:psd_formula}.
\end{proof}

\subsection{Frequency-Dependent Dynamics in Motion Generation}
\label{sec:freq_dynamics}

The result from Theorem \ref{theorem:psd_simplified} helps elucidate the spectral dynamics during diffusion-based motion generation. Natural motion signals $\mathbf{m}_0$ typically exhibit a low-pass characteristic, meaning their power spectral density (PSD) decays with frequency: $|\hat{\mathbf{m}}_0(\omega)|^2 \propto |\omega|^{-\alpha}$ for some $\alpha > 0$.

\paragraph{\textbf{Signal-to-Noise Ratio (SNR)}}
During the forward diffusion process, the signal-to-noise ratio (SNR) at frequency $\omega$ and time $t$ is defined as:
\begin{equation}
    \text{SNR}(\omega, t) = 
    \frac{|\hat{\mathbf{m}}_0(\omega)|^2}{\int_0^t g^2(s)\,\mathrm{d}s},
    \label{eq:snr}
\end{equation}
where $g(t)$ denotes the noise schedule. For a given SNR threshold $\gamma > 0$, we define $t_\gamma(\omega)$ as the earliest time at which the SNR at frequency $\omega$ drops to $\gamma$:
\begin{equation}
    \text{SNR}(\omega, t_\gamma(\omega)) = \gamma
    \quad \Longrightarrow \quad
    \int_0^{t_\gamma(\omega)} g^2(s)\,\mathrm{d}s = \frac{|\hat{\mathbf{m}}_0(\omega)|^2}{\gamma}.
    \label{eq:snr_threshold}
\end{equation}

\paragraph{\ref{form:human_eval}}
Assume the noise schedule is constant, i.e., $g(t) = \sigma$ for some $\sigma > 0$. Then $g^2(s) = \sigma^2$ and the integral simplifies to:

\begin{equation}
    \int_0^{t_\gamma(\omega)} \sigma^2\,\mathrm{d}s = \sigma^2 t_\gamma(\omega).
\end{equation}

Substituting into Equation~\eqref{eq:snr_threshold} gives:

\begin{equation}
    \sigma^2 t_\gamma(\omega) = \frac{|\hat{\mathbf{m}}_0(\omega)|^2}{\gamma}
    \quad \Longrightarrow \quad
    t_\gamma(\omega) = \frac{|\hat{\mathbf{m}}_0(\omega)|^2}{\gamma \sigma^2}.
    \label{eq:t_gamma_constant}
\end{equation}

Using the assumption $|\hat{\mathbf{m}}_0(\omega)|^2 = K |\omega|^{-\alpha}$ for some constant $K > 0$, we obtain:

\begin{equation}
    t_\gamma(\omega) = \frac{K}{\sigma^2 \gamma} |\omega|^{-\alpha}.
    \label{eq:t_gamma_final}
\end{equation}

\paragraph{\textbf{Interpretation.}}
Equation~\eqref{eq:t_gamma_final} reveals several key properties:
As the frequency $\omega$ increases, the corresponding time $t_\gamma(\omega)$ at which the signal-to-noise ratio (SNR) reaches the threshold $\gamma$ decreases. This implies that high-frequency components ($\omega_H$) hit the SNR threshold earlier and thus become corrupted by noise sooner during the forward diffusion process. In contrast, low-frequency components ($\omega_L$) maintain a higher SNR for a longer period, allowing their structural information to be preserved deeper into the diffusion process.

\subsubsection*{Forward Process (Corruption)}
As time $t$ increases, the noise energy $\int_0^t g^2(s)\,\mathrm{d}s$ accumulates, decreasing the SNR across all frequencies. Since high-frequency components exhibit lower spectral energy $|\hat{\mathbf{m}}_0(\omega)|^2$, they reach the SNR threshold $\gamma$ sooner, implying that fine-grained motion details are lost earlier during forward diffusion.

\subsubsection*{Reverse Process (Generation/Denoising)}
The reverse process begins at a high-noise state (large $t$), where low SNR conditions prevail. During denoising, the model first reconstructs low-frequency components, which retain relatively higher SNR and thus guide the recovery of the coarse semantic structure. As time decreases, the effective noise level drops and the SNR improves across all frequencies, allowing the model to progressively refine the motion with higher-frequency details.

This analysis substantiates prior empirical findings~\cite{hu2024ella} that diffusion models tend to generate coarse, low-frequency structure early in the reverse process, followed by high-frequency refinements as the process evolves.

\section{More Visualization Comparison}
\label{app:more_vis}
\begin{figure}
    \centering
    \includegraphics[width=0.9\linewidth]{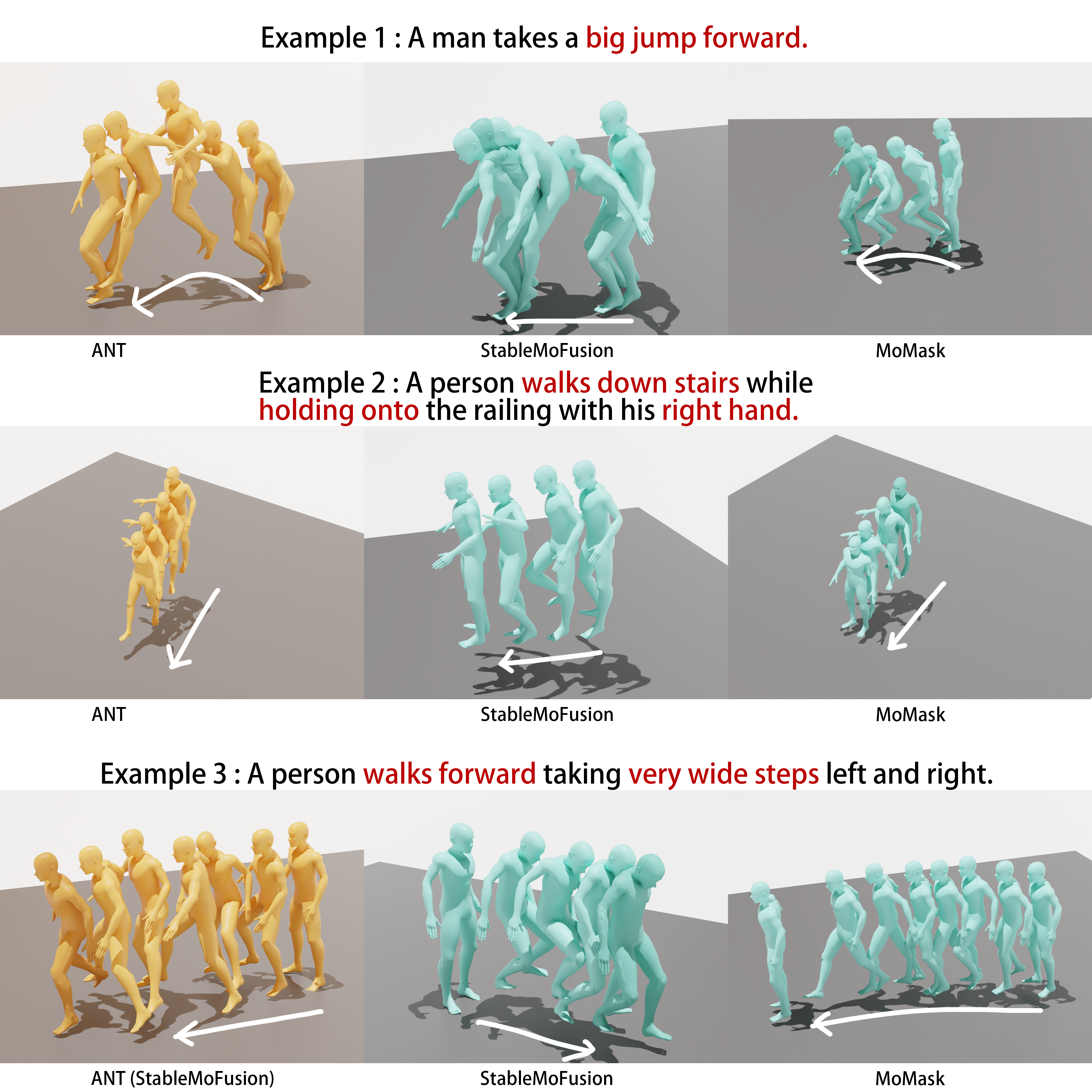}
    \vspace{-0.4cm}
    \caption{Our ANT can be seamlessly plugged into diffusion-based text-to-motion models to generate semantically rich, fine-grained, and naturally smooth motions with high precision.}
    \label{tab:appendixfigure}
    \vspace{-0.6cm}
    
\end{figure}

Figure \ref{tab:appendixfigure} shows the visualization results of ANT and other SOTA models. As can be seen from various examples, ANT outperforms other models in both naturalness and semantic alignment.

\section{DETAILS OF HUMAN EVALUATION}
\label{sec:DETAILS OF HUMAN EVALUATION}
We utilize the Google Form platform to allow 20 individuals to separately fill out 100 different motion sequences pairs for testing, where we have designed two types of questions. The first type involves directly rating the semantic accuracy of generated motion. Motion is presented in GIF format, accompanied by two evaluation options: yes or no. The second type of question pertains to user preferences between our model and a baseline model. This question aims to obtain a comparison of our method and the original method from the user's perspective regarding motion generation accuracy. Our questionnaire takes the form \ref{form:human_eval}: 
\definecolor{myTealLight}{rgb}{0.2, 0.8, 0.8} 
\begin{tcolorbox}[colback=gray!20,arc=2mm]
\label{form:human_eval}
\color{red}\textbf{[Question1]: }\color{black}Is <motion1.gif> semantically accurate?

\begin{enumerate}
\item Yes
\item No
\end{enumerate}

\color{red}\textbf{[Question2]: }\color{black}Is <motion2.gif> semantically accurate?

\begin{enumerate}
\item Yes
\item No
\end{enumerate}

\color{red}\textbf{[Question2]: }\color{black}Which motion result do you think is generated better?
\begin{enumerate}
\item The first one
\item The second one
\end{enumerate}
\end{tcolorbox}

\section{Discussion of Evaluation Metrics}
Previous work has primarily focused on two metrics: FID and R-Precision. FID measures the distance between distributions based on a normality assumption, thereby evaluating the generation quality. However, previous studies \cite{meng2024rethinkingdiffusiontextdrivenhuman, 10.1145/3664647.3681034} have shown that an extremely low FID often contradicts human subjective preferences and is no longer reliable. Considering that the current FID for text-to-motion tasks is already much lower than the average level of text-to-image (T2I) tasks, in this paper, FID is regarded as the second most important reference metric, after R-Precision.

\section{Pseudo-code}
Algorithm \ref{alg:ANT} shows ANT's Pseudo-code
\section{Hyperparameter search result}

\begin{table}[t]
\centering
\resizebox{\columnwidth}{!}{%
   \label{tab:grid}
    \begin{tabular}{|c|c|c|c|c|c|}
        \hline
        \textbf{min $\backslash$ max} & \textbf{2.5} & \textbf{3.0} & \textbf{3.5} & \textbf{4.0} & \textbf{4.5} \\
        \hline
        0.5 & 0.5536/0.1034 & 0.5552/0.0979 & 0.5511/0.0975 & 0.5490/0.0954 & 0.5553/0.0924 \\
        \hline
        1.0 & 0.5502/0.0945 & 0.5546/0.0994 & 0.5578/0.0967 & 0.5562/0.0960 & 0.5509/0.1003 \\
        \hline
        1.5 & 0.5544/0.0974 & \textbf{0.5623/0.0975} & 0.5550/0.0993 & 0.5529/0.0996 & 0.5600/0.0926 \\
        \hline
        2.0 & 0.5528/0.1033 & 0.5494/0.0954 & 0.5575/0.0979 & 0.5535/0.0931 & 0.5498/0.0936 \\
        \hline
        2.5 & 0.5566/0.0975 & 0.5613/0.0992 & 0.5517/0.0984 & 0.5513/0.0943 & 0.5549/0.0962 \\
        \hline
    \end{tabular}
}
\caption{Performance Metrics (Top1/FID) organized in a grid with hyperparameter 
 $\omega_{\min}$ value as rows and hyperparameter $\omega_{\max}$ values as columns (best Top1 highlighted)}
\label{table:hyper}
\vspace{-10pt}
\end{table}

In our study, we conducted a grid search on the validation set to determine the most appropriate values for the hyperparameters $\omega_{\min}$ and $\omega_{\max}$. Specifically, we selected five candidate values for each of $\omega_{\min}$ and $\omega_{\max}$, resulting in a total of 25 distinct parameter combinations. These five groups of hyperparameters were chosen by extending the conventional ranges used in previous studies, allowing us to explore a broader spectrum of possible values.
For each combination, we performed tests five times to improve the reliability and statistical significance of our observations.

After completing our grid search methodology, we first analyzed the relationship between the FID and top-k metrics over various hyperparameter configurations. At higher ranges of the parameters, we observed some level of inverse relationship between FID and top-k accuracy. Specifically, settings that yielded lower FID scores tended to correspond with reduced top-k performance, and vice versa. 
Furthermore, our analysis revealed that, for most of the hyperparameter values (except for the extremely high or low extremes), the overall impact on the results is not significantly sensitive. This insensitivity in the mid-range values suggests that our method demonstrates robustness with respect to these hyperparameter variations.
Based on these observations, we carefully selected our hyperparameter to achieve an optimal balance between FID and top-k performance. Our final choice reflects a compromise that leverages the robust region of the parameter space while mitigating the adverse effects observed at the extremes.

\begin{algorithm}[H]
\caption{ANT Diffusion Process for Text-to-Motion Generation}
\label{alg:ANT}
\begin{algorithmic}[1]
\REQUIRE 

\begin{itemize}
    \item Text prompt $\mathcal{T}$
    \item text encoder (e.g., T5 or CLIP) $Encoder$
    \item Diffusion denoising network $G_\theta$
    \item Diffusion timesteps $T$ with noise schedule $\{\alpha_t\}_{t=1}^T$
    \item Guidance parameters: $\omega_{\max}$, $\omega_{\min}$, decay function $\phi(t)$, threshold time $t_{\text{th}}$
\end{itemize}
\ENSURE Generated motion sequence $x_0$

\medskip
\STATE \textbf{Initialization:}
\STATE $c \gets \text{Encoder}(\mathcal{T})$ \quad \COMMENT{Compute text embedding}
\STATE $x_t \sim \mathcal{N}(0, I)$ \quad \COMMENT{Sample initial noise}
\medskip

\FOR{$t = T$ \textbf{to} $1$ \textbf{by} $-1$}
    \STATE \textbf{Compute Timestep-Specific Features:}
    \STATE $c_t \gets \text{STA}(c, t)$ 
    \quad \COMMENT{Fuse text embedding $e$ with temporal information}
    \medskip
    \STATE \textbf{Denoising Predictions:}
    \STATE $\epsilon_{\text{cond}} \gets G_\theta(x_t, t, c_t)$ \quad \COMMENT{Conditional prediction}
    \STATE $\epsilon_{\text{uncond}} \gets G_\theta(x_t, t, \emptyset)$ \quad \COMMENT{Unconditional prediction}
    \medskip
    \STATE \textbf{Adaptive Guidance Scale:}
    \STATE Compute $\omega_t = \omega_{\min} + \phi(t) \cdot (\omega_{\max} - \omega_{\min})$
    \quad \COMMENT{e.g., using a cosine schedule}
    \medskip
    \STATE \textbf{Guidance and Update:}
    \IF{$t > t_{\text{th}}$} 
        \STATE $\epsilon_t \gets \epsilon_{\text{uncond}}$ 
        \quad \COMMENT{Late stage: use unconditional branch only}
    \ELSE
        \STATE $\epsilon_t \gets \epsilon_{\text{uncond}} + \omega_t \cdot (\epsilon_{\text{cond}} - \epsilon_{\text{uncond}})$
    \ENDIF
    \medskip
    \STATE Update $x_{t-1}$ using the diffusion sampler (e.g., DPM-Solver++):
    \[
    x_{t-1} \gets \text{Update}(x_t, \epsilon_t, t)
    \]
\ENDFOR
\medskip
\RETURN $x_0$
\end{algorithmic}
\end{algorithm}

\section{Deeper Analysis of the STA Module's Internal Behavior}
\label{sec:appendix_sta_analysis}

To provide a deeper, empirical validation of our proposed Semantic Temporally Adaptive (STA) module, this section delves into its internal working dynamics. We aim to visually substantiate our core hypothesis: the STA module facilitates a two-stage denoising process by dynamically modulating the influence of textual semantics over the diffusion timestep $t$ . We present two complementary pieces of evidence: (1) the temporal dynamics of semantic attention within the cross-attention layers, and (2) the evolution of the AdaLayerNorm modulation parameters that control the strength of semantic injection.

\subsection{Temporal Attention Dynamics}

\begin{figure}
    \centering
    \includegraphics[width=0.9\linewidth]{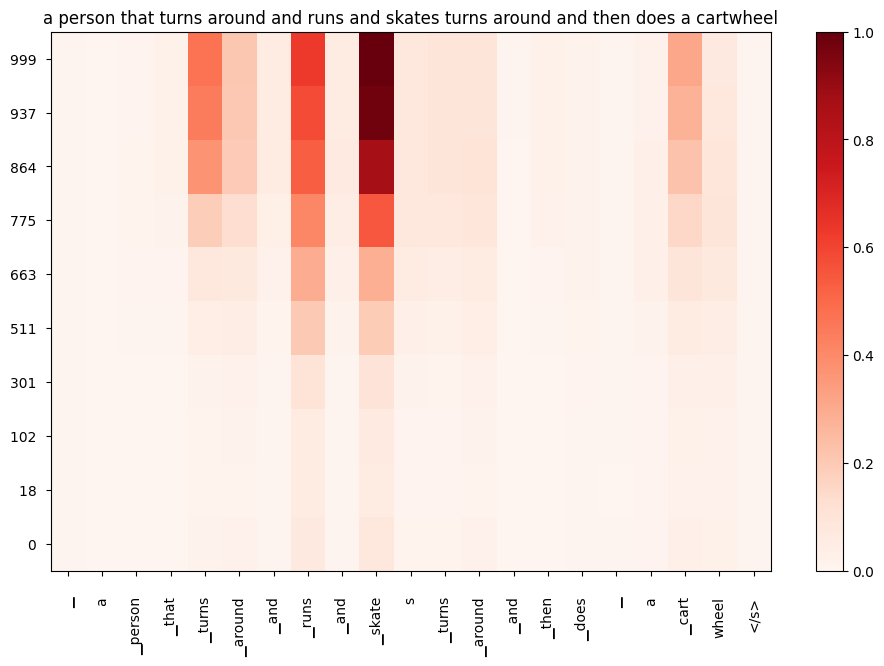}
    \caption{Visualization of attention scores for a sample prompt across diffusion timesteps. The y-axis represents the timestep $t$ from 999 (early, high-noise stage) down to 0 (late, low-noise stage). The x-axis shows the tokens of the input text. Darker shades of red indicate higher attention scores, signifying stronger focus from the model.}
    \label{fig:appendix_attention}
\end{figure}

To understand what semantic information the model prioritizes at different stages, we visualize the cross-attention scores between the motion features and the input text tokens across the denoising timesteps. As illustrated in Figure \ref{fig:appendix_attention}, we analyze the attention patterns for the prompt: ``a person that turns around and runs and skates turns around and then does a cartwheel.''

Our analysis reveals a distinct two-stage behavior:
\begin{itemize}
    \item \textbf{Early Stage (Semantic Planning):} During the initial phase of denoising (e.g., for $t > 700$), the model's attention is strongly concentrated on key tokens that define the motion's high-level structure and core actions. For instance, the words \texttt{runs}, \texttt{skates}, and \texttt{cartwheel} receive substantial attention. This observation aligns perfectly with our concept of a "low-frequency semantic planning" phase, where the model establishes the foundational blueprint of the motion sequence based on global textual cues.
    \item \textbf{Late Stage (Detail Refinement):} As the denoising process progresses into its later stages (e.g., $t < 500$), the attention paid to these high-level semantic tokens rapidly decays, with the heatmap becoming significantly lighter. This indicates a functional shift. Once the core motion structure is established, the model reduces its reliance on explicit semantic guidance and transitions to the "high-frequency detail refinement" phase, where it focuses on generating natural, smooth, and coherent transitions between the established keyframes.
\end{itemize}
\begin{table*}[htbp]
\centering
\resizebox{\textwidth}{!}{%
\begin{tabular}{llcccccc}
\toprule
\textbf{Method} & Venue & \textbf{FID $\downarrow$} & \multicolumn{3}{c}{\textbf{R-Precision $\uparrow$}} & \textbf{Diversity $\rightarrow$} & \textbf{Multimodality $\uparrow$}\\ \cmidrule{4-6}
 & &  & \textbf{Top1} & \textbf{Top2} & \textbf{Top3} &  \\ \midrule
Real  &  & $0.006^{\pm0.003}$ & $0.335^{\pm0.004}$ & $0.513^{\pm0.005}$ & $0.628^{\pm0.002}$ & $10.098^{\pm0.102}$ & - \\
MLD \cite{chen2023executingcommandsmotiondiffusion} & CVPR 2023   & $0.628^{\pm0.038}$ & $0.293^{\pm0.004}$ & $0.459^{\pm0.003}$ & $0.568^{\pm0.004}$ & $9.741^{\pm0.093}$ & $3.035^{\pm0.138}$ \\
T2M \cite{Guo_2022_CVPR} & CVPR 2022 & $1.898^{\pm0.059}$ & $0.252^{\pm0.006}$ & $0.406^{\pm0.005}$ & $0.508^{\pm0.006}$ & $8.975^{\pm0.113}$ & \textcolor{blue}{$4.470^{\pm0.112}$}\\

MoMask \cite{guo2023momask} & CVPR 2024 & $0.383^{\pm0.018}$ & $0.301^{\pm0.005}$ & $0.481^{\pm0.004}$ & $0.597^{\pm0.005}$ & $9.689^{\pm0.092}$ & $1.968^{\pm0.049}$\\
T2M-GPT \cite{t2mgpt} & CVPR 2023 & $0.177^{\pm0.016}$ & $0.353^{\pm0.005}$ & $0.545^{\pm0.006}$ & $0.663^{\pm0.005}$ & \textcolor{blue}{$10.128^{\pm0.132}$} & $1.798^{\pm0.041}$\\
MotionGPT \cite{jiang2023motiongpthumanmotionforeign} & NeurIPS 2023 & $0.267^{\pm0.017}$ & $0.306^{\pm0.004}$ & $0.486^{\pm0.006}$ & $0.605^{\pm0.006}$ & $9.357^{\pm0.133}$ & $2.210^{\pm0.137}$\\
MMM \cite{pinyoanuntapong2024mmmgenerativemaskedmotion}& CVPR 2024 & \textcolor{blue}{$0.151^{\pm0.013}$} & \textcolor{red}{$0.353^{\pm0.004}$} & \textcolor{red}{$0.545^{\pm0.004}$} & \textcolor{red}{$0.667^{\pm0.005}$}& \textcolor{red}{$10.091^{\pm0.086}$} & $0.757^{\pm0.042}$\\

MDM \cite{mdm} & ICLR 2023   & $9.467^{\pm0.217}$ & $0.049^{\pm0.003}$ & $0.098^{\pm0.005}$ & $0.148^{\pm0.005}$ & $7.608^{\pm0.100}$ & \textcolor{red}{$5.682^{\pm0.203}$} \\

StableMoFusion \cite{huang2024stablemofusionrobustefficientdiffusionbased} & ACM MM 2024 & $0.460^{\pm0.003}$ & $0.312^{\pm0.004}$ & $0.494^{\pm0.004}$ & $0.607^{\pm0.005}$ & $9.546^{\pm0.079}$ & $2.157^{\pm0.044}$\\

ANT (StableMoFusion) & - & \textcolor{red}{$0.149^{\pm0.011}$} & \textcolor{blue}{$0.347^{\pm0.005}$} & \textcolor{blue}{$0.541^{\pm0.005}$} & \textcolor{blue}{$0.666^{\pm0.006}$} & $10.034^{\pm0.065}$ & $2.094^{\pm0.059}$\\ 
\bottomrule
\end{tabular}%
}
\caption{Evaluation metrics for CMP dataset. $\pm$ indicates a 95\% confidence interval. \textcolor{red}{Red} and \textcolor{blue}{Blue} indicate the best and the second best result. The right arrow $\rightarrow$ means the closer to real motion the better. Red and Blue indicate the best and the second best result.}
\label{cmp}
\vspace{-0.7cm}
\end{table*}

\subsection{Modulation Strength via AdaLayerNorm Parameters}

To provide a more mechanistic understanding of how the STA module controls semantic influence, we analyze the behavior of its core components. The AdaLayerNorm layer modulates motion features using a time-dependent scale parameter ($\alpha$) and shift parameter ($\beta$), which are derived from the textual condition. The magnitudes of these parameters directly govern the strength of semantic conditioning.

\begin{figure}[h!]
    \centering
    \includegraphics[width=0.9\linewidth]{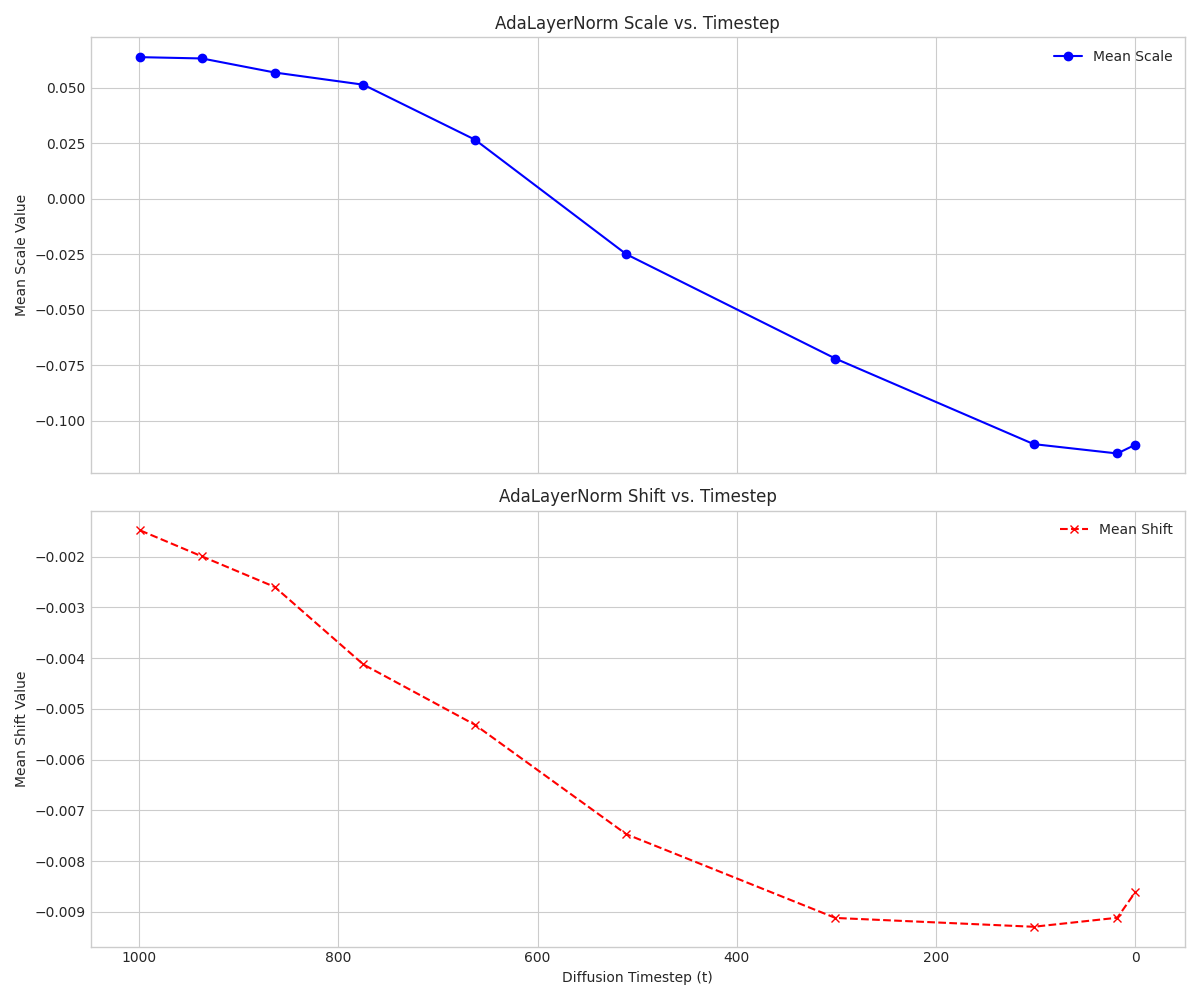}
    \caption{Evolution of the mean \texttt{scale} ($\alpha$, top) and \texttt{shift} ($\beta$, bottom) parameters of the AdaLayerNorm layer within the STA module. The parameters are plotted against the diffusion timestep $t$. The decreasing trend in their values indicates a systematic weakening of semantic modulation strength.}
    \label{fig:appendix_adaln_params}
\end{figure}

Figure \ref{fig:appendix_adaln_params} illustrates the evolution of the mean $\alpha$ and $\beta$ values across all denoising timesteps.

The plots show a clear and systematic decrease in the values of both $\alpha$ and $\beta$ as the timestep $t$ decreases from 1000 to 0. This provides direct quantitative evidence that the STA module programmatically and gracefully weakens the influence of textual semantics as the denoising process unfolds. This behavior is the explicit mechanism behind the "temporal-semantic reweighting" central to our work. It ensures that semantic guidance is strongest when needed for structural planning in the early stages and is attenuated during the later stages to allow for fine-grained, naturalistic motion refinement.

\subsection{Synthesis}
In summary, these two analyses offer a multi-faceted and cohesive view of the STA module's internal dynamics. The attention heatmap (Figure \ref{fig:appendix_attention}) reveals \textit{what} the model focuses on, demonstrating a shift from high-level concepts to implicit details. Concurrently, the AdaLayerNorm parameter plots (Figure \ref{fig:appendix_adaln_params}) reveal \textit{how strongly} this focus is applied, showing a programmed decay in semantic modulation strength. Together, they provide compelling empirical validation that our STA module successfully implements the intended adaptive, two-stage semantic injection, which is a key contributor to the superior performance and semantic alignment achieved by the ANT model.

\section{More Comparison Experiments}
\label{sec:appendix_more_exp}

To further validate the robustness and generalizability of our proposed ANT model, we provide additional comparison experiments on the Combat Motion Processed (CMP) dataset. This dataset features a distinct motion style compared to the more general HumanML3D and KIT benchmarks, serving as a challenging test case for text-to-motion generation.

\paragraph{CMP Dataset.}
The Combat Motion Processed (CMP) dataset \cite{cmp} is a benchmark with a combat motion style that includes 8,700 motions and 26,100 text descriptions. It serves as a smaller-scale but more challenging evaluation benchmark for text-to-motion models. For our experiments on this dataset, we employ the same setup used for the main benchmarks: the pose representation follows StableMoFusion, motions are augmented through mirroring, and the data is split into training, validation, and testing sets with a ratio of 0.8 : 0.15 : 0.05.


\paragraph{Results on CMP}
Table \ref{cmp} presents the quantitative results on the CMP dataset. As shown, our ANT model significantly outperforms the baseline (StableMoFusion) across all key metrics, including FID and R-Precision. This demonstrates the effectiveness of our adaptive temporal-aware architecture in a specialized and challenging domain. Furthermore, ANT achieves competitive or superior performance when compared to other state-of-the-art methods, highlighting the strong generalizability of our proposed approach beyond common human actions. These results on the CMP dataset further validate that ANT provides a robust and effective enhancement for diffusion-based text-to-motion models.

\section{Evaluation Metrics}

\begin{itemize}[noitemsep,leftmargin=*]

    \item \textbf{Frechet Inception Distance  (FID):} We can evaluate the overall motion quality by measuring the distributional difference between the high-level features of the motions. 
    \item \textbf{R-Precision:} We rank Euclidean distances between a given motion sequence and 32 text descriptions (1 ground-truth and 31 randomly selected mismatched descriptions). We report Top-1, Top-2, and Top-3 accuracy of motion-to-text retrieval. 
    
    \item \textbf{Diversity:} From a set of motions, we randomly sample 300 pairs and compute the average Euclidean distances between them to measure motion diversity.
    \item \textbf{Multimodality:} For one text description, we generate 20 motion sequences forming 10 pairs of motion. We then extract motion features and compute the average Euclidean distances of the pairs. We finally report the average over all the text descriptions.
    \item \textbf{Multimodal Distance (MM-Dist)}. The average Euclidean distances between each text feature and the generated motion feature from this text.
\end{itemize}

\end{document}